\documentclass{article} % For LaTeX2e
\usepackage{iclr2024_conference,times}

\usepackage{amsmath,amsfonts,bm}

\def\eqref#1{equation~\ref{#1}}
\def\1{\bm{1}}

\def\va{{\bm{a}}}

\def\vo{{\bm{o}}}

\def\vz{{\bm{z}}}

\def\mI{{\bm{I}}}

\def\mX{{\bm{X}}}

\DeclareMathAlphabet{\mathsfit}{\encodingdefault}{\sfdefault}{m}{sl}
\SetMathAlphabet{\mathsfit}{bold}{\encodingdefault}{\sfdefault}{bx}{n}

\usepackage[pagebackref=true,bookmarks=true]{hyperref}
\usepackage{url}

\usepackage{amsmath}
\usepackage{amssymb}
\usepackage{amsthm}
\usepackage{bm}
\usepackage{graphicx}
\usepackage[group-separator={,}]{siunitx}
\usepackage{caption}
\usepackage{subcaption}
\usepackage{scalerel}
\usepackage{stackengine,wasysym}
\usepackage{algorithm}
\usepackage{algorithmic}
\usepackage{enumitem}
\usepackage{multirow}
\usepackage{pifont}
\usepackage[flushleft]{threeparttable}

\usepackage{booktabs}       % professional-quality tables
\usepackage{amsfonts}       % blackboard math symbols
\usepackage{nicefrac}       % compact symbols for 1/2, etc.
\usepackage{microtype}      % microtypography
\usepackage{xcolor}         % colors

\usepackage{wrapfig}

\definecolor{mydarkblue}{rgb}{0,0.08,0.45}
\definecolor{mydarkgreen}{RGB}{0, 139, 69}
\hypersetup{
    breaklinks=true,
    colorlinks=true,
    urlcolor=mydarkblue,
	citecolor=mydarkblue,
    linkcolor=mydarkblue
}
\definecolor{mycyan}{cmyk}{.3,0,0,0}

\definecolor{input}{RGB}{46, 117, 182}
\definecolor{output}{RGB}{165, 0, 33}

\newcommand{\vth}{\bm{\theta}}
\newcommand{\veps}{\bm{\epsilon}}

\newcommand{\modify}[1]{{{#1}}}

\title{RDT-1B: a Diffusion Foundation Model for \\ Bimanual Manipulation}
\iclrfinalcopy

\author{%
  Songming Liu\thanks{Equal contribution;~~$^\dag$Corresponding authors at dcszj@tsinghua.edu.cn},~
  Lingxuan Wu$^{*}$,~ 
  Bangguo Li,~
  Hengkai Tan,~
  Huayu Chen,\\
  \textbf{Zhengyi Wang},~
  \textbf{Ke Xu},~
  \textbf{Hang Su}$^{\dag}$, ~
  \textbf{Jun Zhu}$^\dag$ 
   \\
   $^{1}$Department of Computer Science \& Technology, Institute for AI, BNRist Center,\\
   Tsinghua-Bosch Joint ML Center, THBI Lab, Tsinghua University\\
}

\begin{document}

\maketitle

\begin{abstract}
Bimanual manipulation is essential in robotics, yet developing foundation models is extremely challenging due to the inherent complexity of coordinating two robot arms (leading to multi-modal action distributions) and the scarcity of training data. In this paper, we present the Robotics Diffusion Transformer (RDT), a pioneering diffusion foundation model for bimanual manipulation. RDT builds on diffusion models to effectively represent multi-modality, with innovative designs of a scalable Transformer to deal with the heterogeneity of multi-modal inputs and to capture the nonlinearity and high frequency of robotic data. To address data scarcity, we further introduce a Physically Interpretable Unified Action Space, which can unify the action representations of various robots while preserving the physical meanings of original actions, facilitating learning transferrable physical knowledge. With these designs, we managed to pre-train RDT on the largest collection of multi-robot datasets to date and scaled it up to $1.2$B parameters, which is the largest diffusion-based foundation model for robotic manipulation. We finally fine-tuned RDT on a self-created multi-task bimanual dataset with over $6$K+ episodes to refine its manipulation capabilities. Experiments on real robots demonstrate that RDT significantly outperforms existing methods. It exhibits zero-shot generalization to unseen objects and scenes, understands and follows language instructions, learns new skills with just 1$\sim$5 demonstrations, and effectively handles complex, dexterous tasks. We refer to the \href{https://rdt-robotics.github.io/rdt-robotics/}{project page} for the code and videos.

\end{abstract}

%is a fundamental task %in obotics~\citep{mirrazavi2017dynamical,rakita2019shared,grannen2023learning} and 
\section{Introduction}
Bimanual manipulation is essential for robots to accomplish real-world tasks~\citep{edsinger2007two}. For practical applications, a useful manipulation policy should be able to generalize to unseen scenarios, such as unseen objects and scenes. However, current approaches either depend on task-specific primitives~\citep{mirrazavi2017dynamical,rakita2019shared,grannen2023learning} or are limited to small-scale model, data and simple tasks~\citep{krebs2021kit,franzese2023interactive,grannen2023stabilize,zhao2023learning,grotz2024peract2,liu2024voxact}, thereby exhibiting only narrow generalization and failing in complex tasks. Following the success in natural language processing~\citep{achiam2023gpt,touvron2023llama} and computer vision~\citep{radford2021learning,kirillov2023segment}, one promising direction to enable generalizable behaviors is to develop a foundation model through imitation learning on large-scale datasets.

Developing bimanual manipulation foundation models confronts the dual challenges of data scarcity and architectural limitations. The prohibitive costs of dual-arm systems create severe data scarcity~\citep{sharma2018multiple,padalkar2023open}, fundamentally conflicting with the data-hungry nature of foundation models. Inspired by recent attempts in unimanual manipulation~\citep{brohan2023rt,kim2024openvla}, we mitigate this through cross-robot pretraining: leveraging multi-robot datasets for pre-training followed by target-robot fine-tuning, amplifying data volume by three orders of magnitude to extract transferable physical priors. However, two interconnected technical barriers emerge. First, the doubled action space induces multi-modal action distributions~\citep{li2006optimal,jia2024towards} (see Fig.~\ref{fig:toy example b} for an illustrative example) that demand \emph{expressiveness} capability beyond current methods~\citep{zhao2023learning,brohan2023rt,kim2024openvla}, while simultaneously requiring \emph{scalabilty} for stable large-scale training on multimodal data (text, vision, actions). Beyond architectural constraints, physical and action space variations across robots introduce data heterogeneity that risks negative transfer~\citep{pan2009survey}. Existing solutions either discard robots with structural inconsistencies or retain only cross-robot invariant features~\citep{brohan2023rt,team2023octo,shah2023gnm}, sacrificing valuable data diversity essential for generalization.

\begin{figure}[t]\vspace{-3ex}
    \centering
    \includegraphics[width=\textwidth]{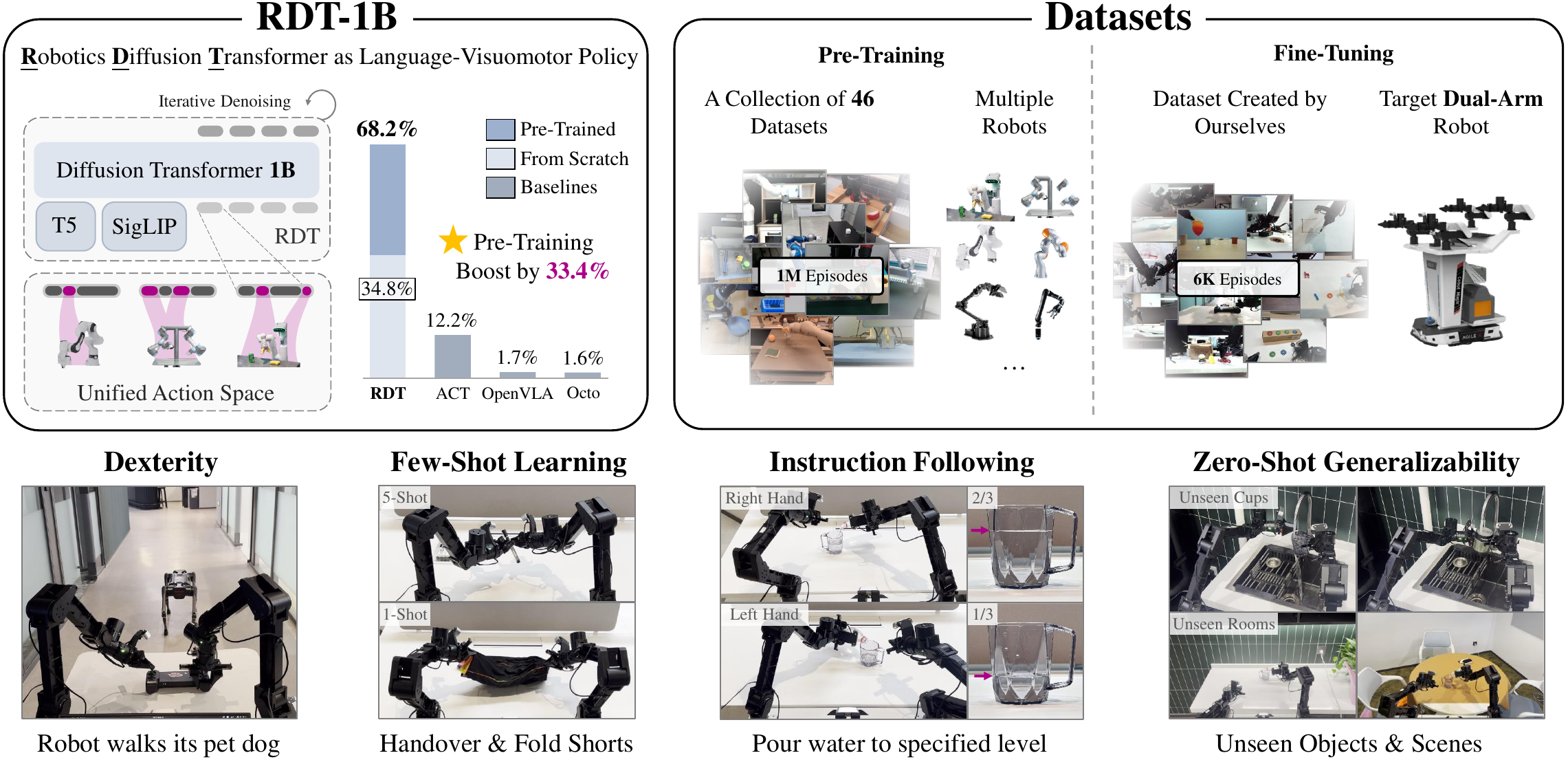}
    % \fbox{\rule{0pt}{1.8in} \rule{0.95\linewidth}{0pt}}
    \vspace{-3ex}
    \caption{\textbf{Overview of Robotics Diffusion Transformer with 1B-Parameters  (RDT-1B)}, a language-conditioned visuomotor policy for bimanual manipulation,% pre-trained on $1$M+ multi-robot episodes and fine-tuned on $6$K+ self-created bimanual episodes. RDT achieves 
    with state-of-the-art generalizability to unseen scenarios (See App.~\ref{app:exp} for metric calculation details). %Code and a Demo video are provided in the supplementary materials.
    }
    \vspace{-3ex}
    \label{fig:head-demo}
\end{figure}

In this paper, we introduce the \emph{Robotics Diffusion Transformer (RDT)}, the largest bimanual manipulation foundation model with strong generalizability. RDT employs diffusion transformer (DiT) as \modify{its scalable backbone~\citep{peebles2023scalable}}, with special designs for language-conditioned bimanual manipulation. For expressiveness, RDT excels in capturing the full modalities of bimanual actions from massive data by using the capacity of diffusion models to represent complex distributions~\citep{sohn2015learning,ho2020denoising}. For scalability, we harness the Transformer backbone and carefully design the multi-modal encoding to eliminate the heterogeneity of various modalities. \modify{Moreover, robotic data is differed significantly from images and videos with temporal and spatial continuity~\citep{chen2019capturing,liang2022self}.} To characterize its inherent nonlinear dynamics~\citep{de2012theory}, high-frequency changes~\citep{team2023octo}, and the unstable numerical range, we make important modifications to the original DiT structure, including MLP decoding, improved normalization, and alternate injection of conditions (see Fig.~\ref{fig:necessity} for their importance). To further enable training RDT on heterogeneous data, we propose the \emph{Physically Interpretable Unified Action Space}, a unified action format for various robots with gripper arms. This innovative format mitigates potential conflicts between different robots while retaining the physical meanings of the original actions, which can promote the model to learn generalizable physical knowledge across diverse robotic datasets.

With the above designs, we managed to pre-train the RDT model on the largest collection of multi-robot datasets to date~\citep{padalkar2023open,walke2023bridgedata,fang2023rh20t,RoboHive} and scale it up to $1.2$B parameters, which is the largest diffusion-based pre-trained model for robotic manipulation. To further enhance its bimanual manipulation capabilities, we fine-tuned the RDT on a self-collected multi-task bimanual dataset comprising over $6$K+ trajectories, which is one of the most extensive bimanual datasets. In our experiments, we have comprehensively evaluated RDT against strong baselines in both bimanual manipulation and robotic foundation models. Results show that RDT achieves state-of-the-art performance, outperforming baselines by achieving an improvement of $56\%$ in success rates across a wide spectrum of challenging tasks. In particular, RDT has exceptional zero-shot and few-shot ($1\sim 5$ shots) generalizability to unseen objects, scenes, instructions, and even skills. RDT is also capable of accomplishing tasks requiring fine-grained operations, such as controlling a robot dog with a joystick. Finally, ablation studies show that diffusion modeling, large model size, and large data size all contribute to superior performance.

%===============================================================================

\section{Related Work}
\label{sec:related}

{\bf Learning-based Bimanual Manipulation.} 
One substantial challenge in learning a bimanual manipulation policy is the high dimensionality of the action space, which exacerbates the data scarcity~\citep{zollner2004programming,smith2012dual,lioutikov2016learning,stepputtis2022system} and the multi-modal behavior~\citep{colome2018dimensionality,colome2020reinforcement,figueroa2017learning,sharma2018multiple,xie2020deep,franzese2023interactive}. Some works have developed more cost-effective interfaces for data collection~\citep{zhao2023learning,aldaco2024aloha}, but they are limited to specific hardware configurations and still insufficient to bridge the data gap for a generalizable policy. 
Others attempt to reduce data requirements by introducing inductive biases, such as distinguishing two arms for stabilization and functionality~\citep{grannen2023stabilize}, parameterizing movement primitives~\citep{batinica2017compliant,amadio2019exploiting,chitnis2020efficient,franzese2023interactive}, or using voxel representations~\citep{grotz2024peract2,liu2024voxact}. These methods use strong priors or simplified modeling, which successfully reduce the action space, but at the cost of a reduced scope of application and inability to express the multi-modality of bimanual behaviors~\citep{pearce2023imitating}.

% In contrast, our diffusion model can accurately handle multi-modality, which does not rely on strong priors but instead learns generalizable knowledge through massive amounts of multi-robot data.

%  through large-scale pre-training on diverse datasets, similar to their successes in natural language processing~\citep{achiam2023gpt,touvron2023llama} and computer vision~\citep{radford2021learning,kirillov2023segment}. These models have the potential to empower robots to perform a wide array of tasks by learning from extensive data. However, current approaches are constrained by several limitations, such as a focus on unimanual manipulation, the use of small-scale datasets with limited tasks, and models with relatively few parameters~\citep{krebs2021kit,franzese2023interactive,grannen2023stabilize,zhao2023learning,grotz2024peract2,liu2024voxact}. These limitations hinder the development of generalist robotic models capable of handling complex, real-world scenarios.

% A recent trend in robotics works towards training multi-task ``generalist'' models [2, 6, 45–49] on large diverse robot datasets [1, 2, 6, 11, 45, 49–56]
% , ~\citep{padalkar2023open,brohan2022rt,fang2023rh20t}

% 

{\bf Foundation Models for Robotics.} Foundation models have shown immense promise in enabling generalizable behaviors by training multi-task ``generalist'' models~\citep{brohan2022rt,brohan2023rt,team2023octo,kim2024openvla} on large multi-task robot datasets~\citep{padalkar2023open,brohan2022rt,fang2023rh20t}. Most studies adapt large vision-language models to directly predict action~\citep{brohan2022rt,driess2023palm, brohan2023rt,padalkar2023open,kim2024openvla}. While demonstrating generalization to new objects and tasks, they face issues with quantization errors and uncoordinated behaviors~\citep{pearce2023imitating} when applied to bimanual manipulation. \modify{It's largely due to their discretization of action spaces.} To enhance precision, diffusion models have been used for continuous control~\citep{ho2020denoising,chi2023diffusionpolicy,pearce2023imitating,team2023octo}. \citet{team2023octo} pre-train a Transformer-based diffusion policy on a subset of Open X-Embodiment~\citep{padalkar2023open} dataset ($25$ datasets), with up to $93$M parameters.

\section{Problem Formulation and Challenges}
\label{sec:method:overview}
\begin{figure}[t]\vspace{-3ex}
     \centering
     \begin{subfigure}[b]{0.3\textwidth}
         \centering
         \includegraphics[width=\textwidth]{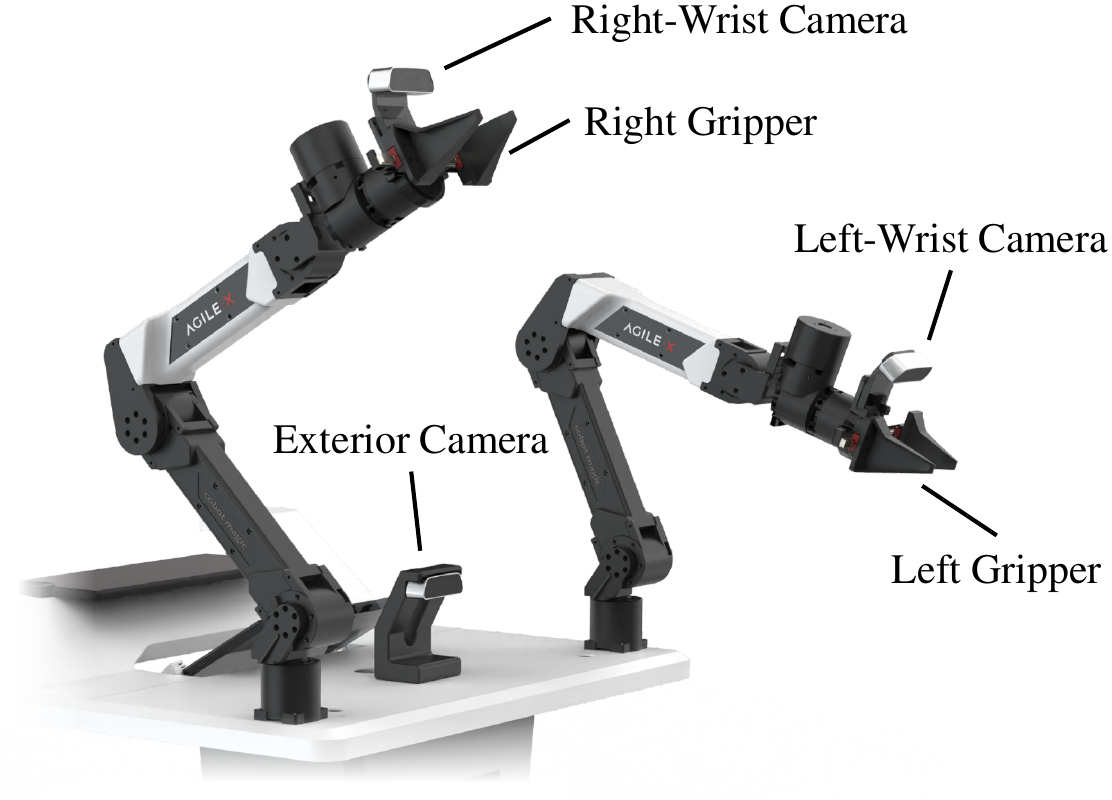}
         \caption{Target dual-arm robot}
         \label{fig:toy example a}
     \end{subfigure}
     \hfill
     \begin{subfigure}[b]{0.55\textwidth}
         \centering
         \includegraphics[width=\textwidth]{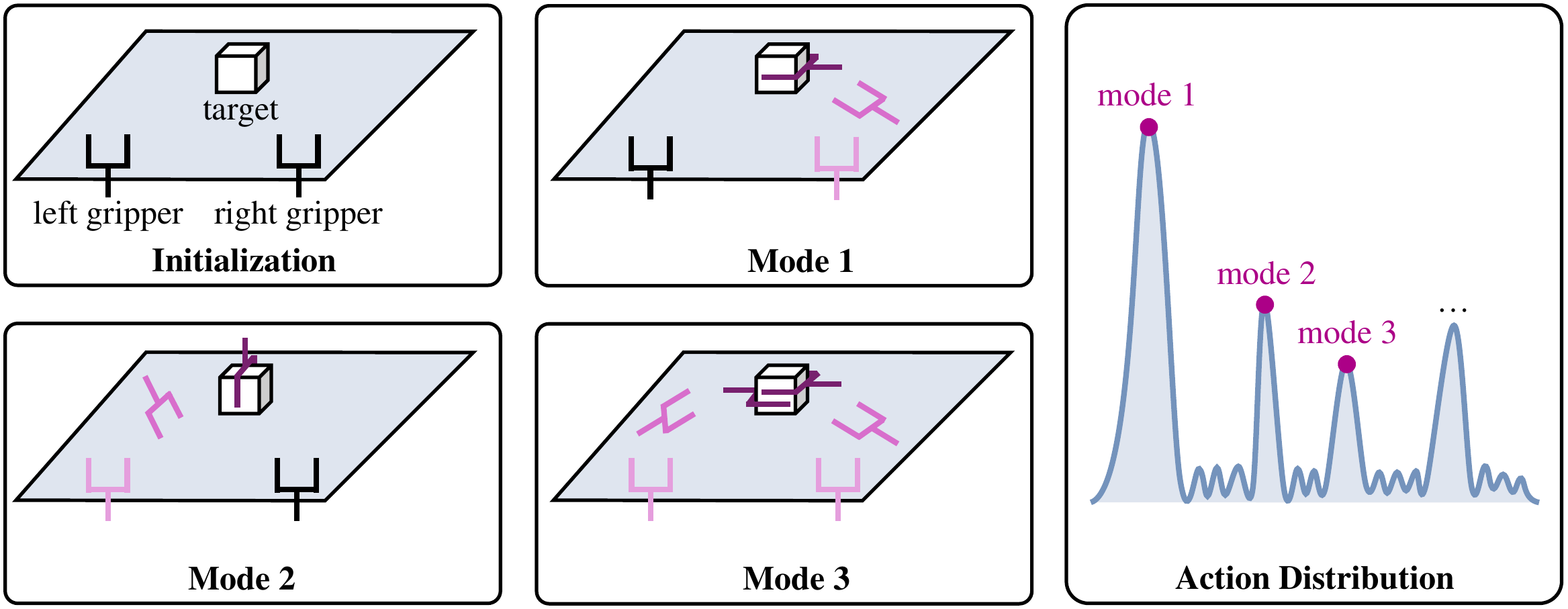}
         \caption{Illustration of multi-modality}
         \label{fig:toy example b}
     \end{subfigure}\vspace{-1ex}
        \caption{\textbf{(a)} Schematic diagram of the ALOHA dual-arm robot. \textbf{(b)} A toy example of grasping a cube. Compared with unimanual manipulation, bimanual manipulation has more possible action modes, leading to stronger multi-modality. Colors from light to dark indicate that time goes forward.}\vspace{-2ex}
        \label{fig:toy example}
\end{figure}

We start by formulating the task %of training a foundation model for bimanual manipulation % to solve language-conditioned bimanual manipulation tasks 
and elaborating on the challenges. To evaluate the model on the hardware, we choose the ALOHA dual-arm robot as our target robot since it is one of the most representative dual-arm robots and is suitable for collecting human demonstration data via teleoperation~\citep{zhao2023learning,fu2024mobile,aldaco2024aloha}. Fig.~\ref{fig:toy example a} shows a schematic diagram of the target robot, which consists of two arms with grippers and three cameras. Note that our setting and foundation model are generic to any dual-arm gripper robot. %, not limited to the target one.

We consider the concrete task of language-conditioned bimanual manipulation with vision, which is fundamental in robotics and has great value in real-world scenarios such as household~\citep{stepputtis2020language,brohan2022rt,zhao2023learning}. Formally, given a language instruction $\ell$, the policy is presented with an observation $\vo_t$ at time $t\in \mathbb{N}^+$; and then it produces an action $\va_t$ to control \emph{two robot arms} to achieve the goal specified by $\ell$. The observation is represented as a triple $\vo_t:= (\mX_{t-T_{\text{img}}+1:t+1}, \vz_t, c)$, where $\mX_{t-T_{\text{img}}+1:t+1}:= (\mX_{t-T_{\text{img}}+1},\dots,\mX_{t})$ is the RGB observation history of size $T_{\text{img}}$, $\vz_t$ is the low-dimensional proprioception of the robot, and $c$ is the control frequency. The action $\va_t$ is usually a subset of the desired proprioception $\vz_{t+1}$\footnote{E.g., $\vz_{t}$ may include the gripper position at time $t$, and $\va_t$ can be the target gripper position at step $t+1$.}.

% Methods from Reinforcement Learning (RL)~\citep{amadio2019exploiting,kataoka2022bi} also fail due to the sim-to-real gap and the difficulty of designing reward functions for general manipulation tasks.
 % or ``wipe the table \underline{from left to right}"

A specific task in bimanual manipulation typically consists of multiple elements: a \emph{skill} (e.g., verbs like ``pick", ``wipe", or ``open"), an \emph{object} (e.g., nouns like ``bottle", ``table", or ``door"), a \emph{scene} (i.e. the environment in which the task takes place), and a \emph{modality} describing how the skill is performed (e.g., adverbials like ``pick the bottle \underline{with the left hand}"). When encountering a new task, a practical policy is required to generalize to unseen\footnote{\emph{unseen} means that a certain element has not appeared in the training data.} elements in the task, which is particularly challenging for previous rule-based methods~\citep{mirrazavi2017dynamical,rakita2019shared,grannen2023learning} as well as learning-based methods that are limited to either small models and data or simple tasks, as discussed in Sec.~\ref{sec:related}. %~\citep{amadio2019exploiting,chitnis2020efficient,kataoka2022bi,chi2023diffusionpolicy}. 

We aim to train a foundation model policy via imitation learning to achieve generalizability. However, the available data for a specific dual-arm robot is particularly scarce ($< 10$K trajectories) due to high hardware costs, far from the common requirement to train a foundation model. To address this, we propose to employ a pre-training and fine-tuning pipeline~\citep{radford2018improving} to take advantage of data from multiple robots by drawing inspiration from recent advances in unimanual manipulation~\citep{team2023octo,padalkar2023open,kim2024openvla}. In this manner, we would expand the data size by three orders of magnitude. Specifically, we first pre-train the model on a large-scale multi-robot dataset $\mathcal{D}_{\text{pre}}$ (mostly single-arm) and then fine-tune on a dataset of the target robot $\mathcal{D}_{\text{ft}}$. We denote the dataset by $\mathcal{D}_{\cdot} = \{ (\ell^{(i)}, \vo^{(i)}_{t}, \va^{(i)}_{t}) \mid 0\le t < T^{(i)}, 1 \le i \le N\}$, where $T^{(i)}$ is the length of the $i$-th trajectory and $N$ is the number of trajectories. Moreover, it is worth emphasizing that our goal is to use multi-robot data to enhance the model's generalizability in bimanual manipulation \emph{rather than} developing a cross-embodiment model for various robots. There are two main challenges to developing such a foundation model with multi-robot data:

\begin{figure}[t]\vspace{-3ex}
    \centering
    \includegraphics[width=.85\textwidth]{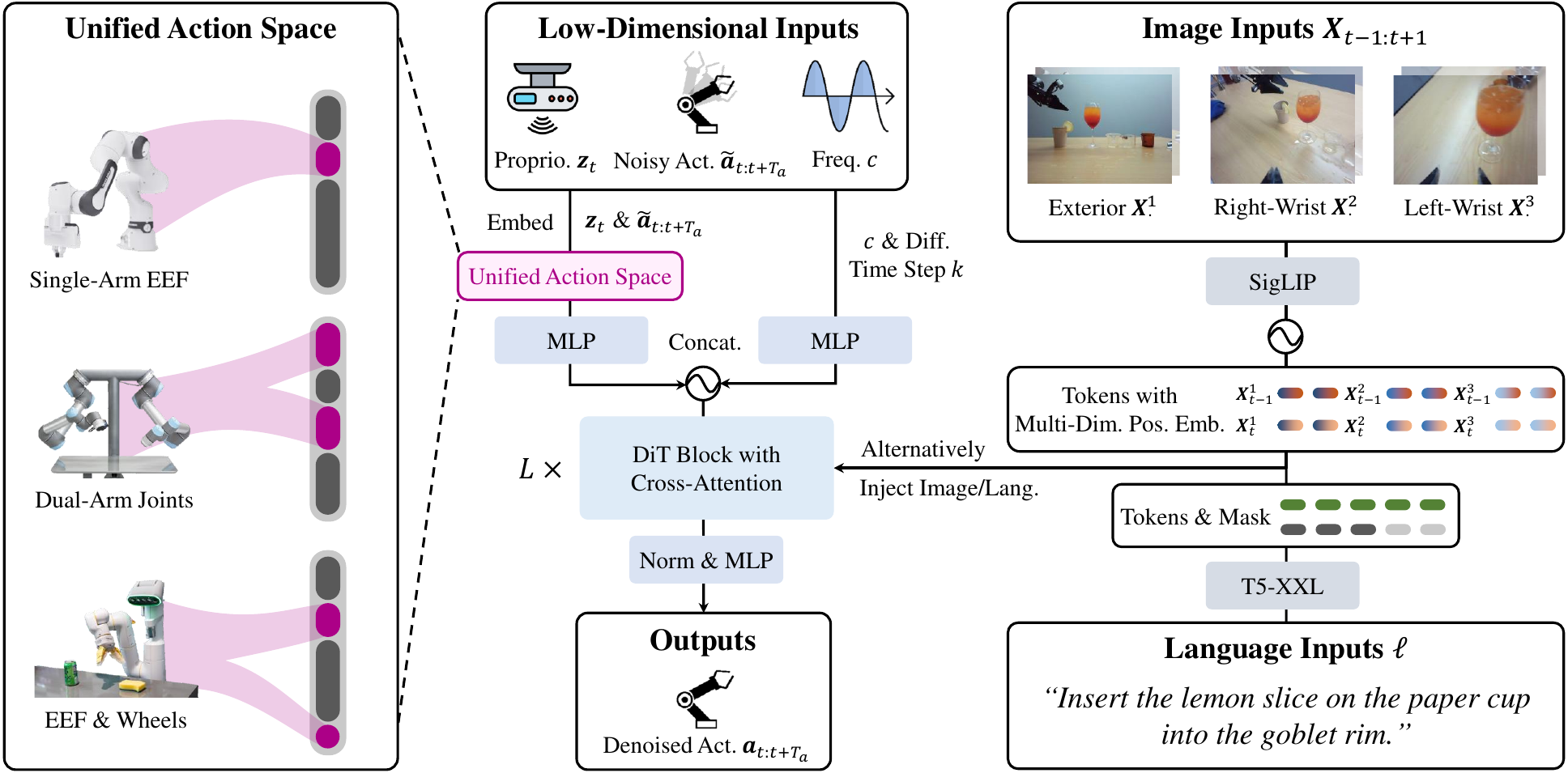}
    % \fbox{\rule{0pt}{2.0in} \rule{0.95\linewidth}{0pt}}
    \caption{\textbf{RDT framework.} Heterogeneous action spaces of various robots are embedded into a unified action space for multi-robot training. \textbf{Inputs:} proprioception $\vz_t$, noisy action chunk $\tilde{\va}_{t:t+T_a}$, control frequency $c$, and diffusion time step $k$, acting as denoising inputs; image inputs ($T_{\text{img}}=2$ and $\mX_{\cdot}=\{ \mX_{\cdot}^1, \mX_{\cdot}^2, \mX_{\cdot}^3 \}$ denotes a set of images from exterior, right-wrist, and left wrist cameras) and language inputs, acting as conditions. \textbf{Outputs:} denoised action chunk $\va_{t:t+T_a}$.}
    \label{fig:framework}\vspace{-2ex}
\end{figure}

{\bf Challenge 1: How to design a powerful architecture?}
A generalizable foundation model necessitates a powerful architecture. This requirement encompasses two primary aspects. Firstly, the architecture must possess sufficient \emph{expressiveness} to capture the multi-modality in the action distribution. Fig.~\ref{fig:toy example b} illustrates a toy example where the robot attempts to grasp a cube. We can see that there are many modes to finish this task, in contrast to unimanual manipulation, where only one robot arm is controlled. When collecting demonstrations, the human operator may randomly pick one of them, leading to multi-modality in the collected action data. Secondly, \emph{scalability} is necessary for such an architecture. As a foundation model, it should effectively process heterogeneous inputs from various modalities (text, images, actions, etc.) while being scalable to train stably on large datasets.

{\bf Challenge 2: How to train on heterogeneous data?}
Training on multi-robot data presents a unique challenge of data heterogeneity. The physical structure and the action space can vary greatly across different robots. Previous attempts either restrict themselves to a subset of robots with similar action spaces~\citep{yang2023polybot,team2023octo,kim2024openvla} or only retain a subset of inputs sharing the same structure~\citep{padalkar2023open,yang2024pushing}, at the cost of losing a lot of information. It remains largely under-addressed on how to train models on such heterogeneous data.

\section{Robotics Diffusion Transformer}\label{sec:model}
We now present Robotics Diffusion Transformer (RDT), as illustrated in Fig.~\ref{fig:framework}. In Sec.~\ref{sec:model:diff}, we present the diffusion model and the corresponding architecture to address Challenge 1. In Sec.~\ref{sec:model:data}, we resolve Challenge 2 by proposing a physically interpretable unified action space to unify various robot action spaces and enable multi-robot pre-training. We also collect a comprehensive multi-task bimanual dataset for fine-tuning to improve the bimanual manipulation capabilities of RDT.

% In Sec.~\ref{sec:model:deploy}, we efficiently deploy our model on the target dual-arm robot for real-time inference.

\subsection{RDT Model}\label{sec:model:diff}

% the imitation learning policy is required to predict an action chunk\footnote{According to \citet{zhao2023learning}, predicting an action sequence rather than a single action is beneficial for mitigating the \textit{compounding error.}} $\va_{t:t+T_a}:=(\va_t, \dots, \va_{t+T_a-1})$ when receiving an observation $\vo_t$ to control two robot arms to achieve the goal; here $T_a$ denotes the chunk size. 

% to model the continuous conditional distribution $p(\va_{t:t+T_a} | \ell, \vo_t)$. 

% \junz{what are key differences from DiT? Make necessary comparison/discussion to elaborate on our novelty: If the problem is new, make it clear what difference from existing formulations; if introducing special designs, make them clear...}

% To represent multi-modality, we resort to an expressive modeling tool, the diffusion model~\citep{diffuser, chen2022offline, pearce2023imitating, chi2023diffusionpolicy}, to be the underlying structure of RDT. We also employ the action chunking technique to alleviate the \emph{compounding error}, as suggested by \citet{zhao2023learning}. That is, we use the diffusion model to express the continuous conditional distribution $p(\va_{t:t+T_a} | \ell, \vo_t)$ instead of $p(\va_{t} | \ell, \vo_t)$, where $\va_{t:t+T_a}:=(\va_t, \dots, \va_{t+T_a-1})$ is an action chunk and $T_a$ denotes the chunk size. 

% {\bf }
\paragraph{Diffusion Modeling.}
Due to multi-modality, given the language instruction $\ell$ and observation $\vo_t$, there may be many possible actions $\va_t$ to proceed with the task. The policy will learn the ``average" of action modes if we model it as a deterministic mapping $(\ell, \vo_t) \mapsto \va_{t}$ and regress the tuples of $(\ell, \vo_t, \va_{t})$ in the training data. This may result in out-of-distribution actions, such as the arithmetic mean of multiple modes, which can be completely infeasible~\citep{pearce2023imitating}. Instead, we choose to model the continuous conditional distribution $p(\va_{t} | \ell, \vo_t)$. 
%Previous works include discretization methods~\citep{brohan2022rt,brohan2023rt,kim2024openvla}, Variational Autoencoder (VAE)~\citep{zhao2023learning,fu2024mobile,aldaco2024aloha}, and diffusion models~\citep{diffuser,chen2022offline,chi2023diffusionpolicy,team2023octo}. 
%In particular, discretization methods divide each dimension of the action space into bins and frame the decision-making as a classification task, thereby suffering from quantization errors and difficulty in accurately approximating the joint distribution of every action dimension~\citep{pearce2023imitating}. VAE assumes that $\va_{t}$ is related to a Gaussian latent variable and sample $\va_{t}$ by Gaussian distribution with learnable mappings, and such assumptions can reduce sampling quality~\citep{dai2019diagnosing}, which is critical for dexterous tasks demanding high action precision. In contrast, 
As discussed in Sec.~\ref{sec:related}, among various approaches, 
diffusion models excel in both expressiveness and sampling quality, but can be slow to \modify{sample high-dimensional data} (e.g., images). Luckily, for our settings, the drawback is minor since that $\va_{t}$ has a much lower dimension than images, which requires only minimal sampling overhead. This has made diffusion models an ideal choice for policy as in \citet{chi2023diffusionpolicy}.

Nevertheless, employing diffusion models for robotic tasks faces unique challenges since the inherent properties of robotic physics quantities (i.e., the action and proprioception) are different from image/video data. Image and video data, while high-dimensional, often exhibit a degree of temporal and spatial continuity~\citep{chen2019capturing,liang2022self}, with changes between frames typically being incremental. In contrast, robotic physics quantities are characterized by its \textit{nonlinear dynamics}~\citep{de2012theory} and the potential for \textit{high-frequency changes} stemming from the physical interactions, such as collision, constraints, and material properties like damping. Moreover, the quantities also feature an \textit{unstable numerical range}, probably due to extreme values caused by unreliable sensors. This underscores the necessity of adapting current diffusion models to effectively capture the instability and nonlinearity of robot data. Next, we will first elaborate on diffusion formulation and then introduce our design of architecture to resolve these challenges.

When making a decision with diffusion policies, we first sample a totally noisy action $\va_t^K \sim \mathcal{N}(\bm{0},\mI)$ and then perform $K\in\mathbb{N}^+$ denoising steps to denoise it to a clean action sample $\va_t^0$ from $p(\va_{t} | \ell, \vo_t)$:
\begin{equation}\label{eq1}
    \va_t^{k-1} = \frac{\sqrt{\bar{\alpha}^{k-1}}\beta^k}{1-\bar{\alpha}^{k}}  \va_t^{0} + \frac{\sqrt{\alpha^{k}}(1-\bar{\alpha}^{k-1})}{1-\bar{\alpha}^{k}} \va_t^{k} + \sigma^k \vz,\quad k=K,\dots,1,
\end{equation}
where $\{ \alpha^k \}_{k=1}^K, \{ \sigma^k \}_{k=1}^K$ are scalar coefficients pre-defined by a noise schedule~\citep{nichol2021improved}. Here, $\beta^k := 1-\alpha^k$, and $\bar{\alpha}^{k-1}:= \prod_{i=1}^{k-1} \alpha^{i}, \vz \sim \mathcal{N}(\bm{0},\mI)$ if $k > 1$, else $\bar{\alpha}^{k-1}=1, \vz = \bm{0}$. However, $\va_t^0$ is intractable before sampling is finished. We opt to use a learnable denoising network $f_{\vth}$ with parameters $\vth$ to estimate the clean sample from a noisy one: $\va_t^{0} \leftarrow f_{\vth}(\ell, \vo_t, \va_{t}^k, k)$. To train such a network, we will minimize the following mean-squared error (MSE) of denoising:
\begin{equation}\label{eq2}
    \mathcal{L}(\vth) := \mathrm{MSE}\left(\va_{t}, f_{\vth}(\ell, \vo_t, \sqrt{\bar{\alpha}^k}\va_{t} + \sqrt{1-\bar{\alpha}^k}\veps, k)\right),
\end{equation}
where $k \sim \mathrm{Uniform}(\{1,\dots, K\})$, $\veps \sim \mathcal{N}(\bm{0},\mI)$, and $(\ell, \vo_t, \va_{t})$ is sampled from our training dataset. Later in this paper, we will denote noisy action inputs by $\tilde{\va}_{t}:= \sqrt{\bar{\alpha}^k}\va_{t} + \sqrt{1-\bar{\alpha}^k}\veps$, in which the superscript of $k$ is dropped for simplicity. Besides, in practice, we prefer to predict a sequence of actions, i.e., an action chunk, in one shot to encourage temporal consistency~\citep{chi2023diffusionpolicy} and to alleviate error accumulation over time by reducing number of decisions in a task~\citep{zhao2023learning}. Specifically, we model $p(\va_{t:t+T_a} | \ell, \vo_t)$, where $\va_{t:t+T_a}:=(\va_t, \dots, \va_{t+T_a-1})$ is an action chunk and $T_a$ denotes the chunk size~\citep{zhao2023learning}. We provide a detailed discussion in App.~\ref{app:ac}.

% Note that different from images or videos, which are high-dimensional yet low-frequency, our physical actions are low-dimensional yet high-frequency.

 % by simulating the reverse diffusion process

% To be specific, during training, we first sample a tuple of $(\ell, \vo_t, \va_{t:t+T_a})$ from $\mathcal{D}_{\cdot}$,\junz{what's $\mathcal{D}_{\cdot}$?}\junz{may need some necessary intro on how diffusion process was adopted to trajectory data and highlight the difference from image/videos? Even need to ellobrate on why diffusion can work here...} a diffusion time step $k\in \mathbb{N}^+$, and a random noise vector $\veps^k$ with variance corresponding to $k$. Then, we train a denoising network\junz{again, with necessary intro to diffusion to get why need this denoising network...} $f_{\vth}$ with parameters $\vth$ to predict the clean action chunk from the noisy one $\tilde{\va}_{t:t+T_a} := \va_{t:t+T_a} + \veps^k$:

% During inference, given $\ell$ and $\vo_t$, we start from a Gaussian noise vector and denoise it to an action chunk sample from $p(\va_{t:t+T_a} | \ell, \vo_t)$ through iteratively applying the trained denoising network $f_{\vth}$. 

% Later in this paper, we will let the model predict $\va_{t:t+T_a}$ instead of $\va_t$, and use $\tilde{\va}_{t:t+T_a}$ to denote a noisy action chunk. 

We now present the design of the architecture, including the encoding of multi-modal inputs and the network structure of $f_{\vth}$, while details are deferred to App.~\ref{app:arc}.

% Although this method is simple, it effectively unifies the representation of different robots. Moreover, 

% In practice, we further model $p(\va_{t:t+T_a} | \ell, \vo_t)$, where $\va_{t:t+T_a}:=(\va_t, \dots, \va_{t+T_a-1})$ is an action chunk and $T_a$ denotes the chunk size, to alleviate \emph{compounding errors}\footnote{\emph{compounding errors} refers to errors that accumulate as the number of historical decisions increases and cause the policy to drift out of the training distribution, reaching hard-to-recover states~\citep{ross2011reduction}.} by reducing the number of decisions in a trajectory.

% RDT processes inputs from various modalities that exhibit significant differences in format and dimensions, 

% {\bf Encoding of Heterogeneous Multi-Modal Inputs.}
\paragraph{Encoding of Heterogeneous Multi-Modal Inputs.}
The heterogeneity of multi-modal inputs is reflected in the structure; that is, the format and number of dimensions of each modality are significantly different. This has posed challenges for mult-modal training. To address this, we encode these diverse modalities into a unified latent space. Below are the encoding methods:
\begin{itemize}[leftmargin=1.0em]
\vspace{-0.5em}
\setlength\itemsep{0em}
    \item \textbf{Low-Dimensional Inputs} are low-dimensional vectors that represent physical quantities of the robot, including the proprioception, the action chunk, and the control frequency. To encode them, we use MLPs (with Fourier features~\citep{tancik2020fourier}), which can effectively capture the \textit{high-frequency changes} in low-dimensional spaces.
    
    % Since they are typically continuous and high-frequency, we use MLPs (with Fourier features~\citep{tancik2020fourier}) to encode these inputs into the latent space. 
    
    % Such continuous encoding can avoid precision loss in contrast to discretized encoding.
    
    % We adopt \textit{in-context conditioning}~\citep{peebles2023scalable,bao2023all} for low-dimensional conditions, including $\vz_t$, $c$ and $k$, to obtain better conditions injection. To be specific, tokens of noisy inputs $\tilde{\va}_{t:t+T_a}$ and low-dimensional conditions are concatenated together in the length direction, resulting in an input token sequence of length $1+T_a+1+1$. Afterward, position embeddings are added to distinguish different modalities and inject temporal information.
    
    % due to the small number of their tokens. In this way, we concatenate all the tokens (i.e., tokens of $\vz_t$, $\tilde{\va}_{t:t+T_a}$, $c$, and $k$) together in the length direction, resulting in an input token sequence of length $1+T_a+1+1$. We also add position embeddings to it to distinguish different modalities and inject the temporal information into the action chunk $\tilde{\va}_{t:t+T_a}$. 
    \item \textbf{Image Inputs} are high-dimensional and contain rich spatial and semantic information. To extract compact representations, we use an image-text-aligned pre-trained vision encoder, SigLIP~\citep{zhai2023sigmoid}. We fix its weights during training to save GPU memory. 

% Widely used
    
    \item \textbf{Language Inputs} are of varying length and highly abstract, posing integration challenges due to their complexity and ambiguity. To encode them, we use a pre-trained Transformer-based language model, T5-XXL~\citep{2020t5}. We also fix its weights during training to save GPU memory.
    
    % When performing cross-attention conditioning, we apply the language attention mask to mask out the pad tokens appended during batching.
\vspace{-0.5em}
\end{itemize}

Besides, heterogeneity also manifests in both cross-modal and intra-modal information density. First, modalities differ inherently in information capacity (e.g., visual inputs typically yield more tokens than text). Second, substantial variance exists within modalities, as seen in robotic perception where exterior cameras capture broader scenes versus the limited views from wrist cameras (Fig.~\ref{fig:framework}). This discrepancy risks shortcut learning: focusing on exterior views while neglecting details from wrist cameras. To promote balanced multimodal integration, we implement stochastic independent masking across modalities during encoding, preventing overreliance on specific inputs.

% model oversimplification through dominant modality bias - for instance, over-indexing on exterior views while neglecting depth cues from wrist cameras. To promote balanced multimodal integration, we implement stochastic independent masking across modalities during encoding, preventing overreliance on specific inputs.

% In addition to structure, heterogeneity features the different amounts of information contained in different inputs. First, data in different modalities contain different amounts of information. For example, images usually contain more information than text, and after encoding, they produce more tokens. Second, different inputs of the same modality may hold very different amounts of information. For example, the exterior camera of a robot has a more global view and contains richer information than the wrist cameras, as shown in the upper right of Fig.~\ref{fig:framework}. In this case, the model may learn a shortcut: only focusing on the exterior view and ignoring the wrist views, thereby losing the ability to perceive depth. To tackle this issue, we randomly and independently mask each multi-modal input with a certain probability during encoding to prevent the model from over-relying on a specific input. 

% We use Transformer~\citep{vaswani2017attention} as the basic architecture of our denoising network $f_{\vth}$ due to its high scalability
% \paragraph{Random Input Mask.}

% These physical quantities are typically continuous, obeying nonlinear physical laws~\citep{de2012theory}. Therefore, improving the approximation ability of nonlinearity

% \newpage

\begin{wrapfigure}{r}{0.3\linewidth}
\centering
 \vspace{-1em}
% \fbox{\rule[-.5cm]{0cm}{4cm} \rule[-.5cm]{4cm}{0cm}}
\begin{minipage}{\linewidth}
    \captionsetup[subfigure]{justification=centering,font=small}
    \centering
    \includegraphics[width=0.9\linewidth]{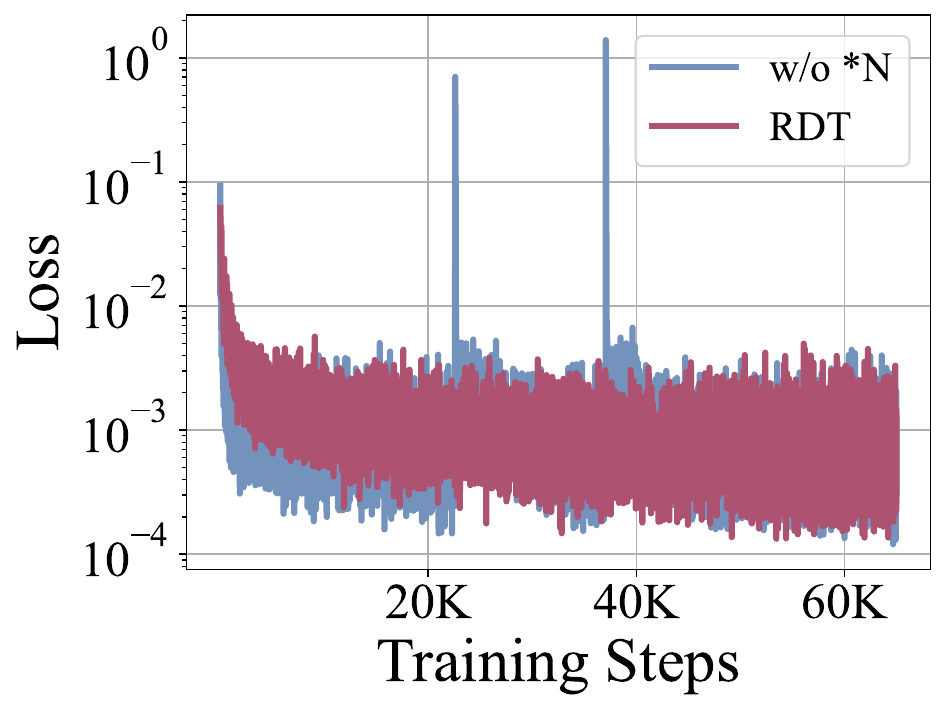}
 \vspace{-0.5ex}
    \subcaption{\modify{Loss w/o QKN \& RMSN}}
    \label{fig:necessity:a}
    % \hfill
   % \par\vfill
    \vspace{1ex}
    \includegraphics[width=0.9\linewidth]{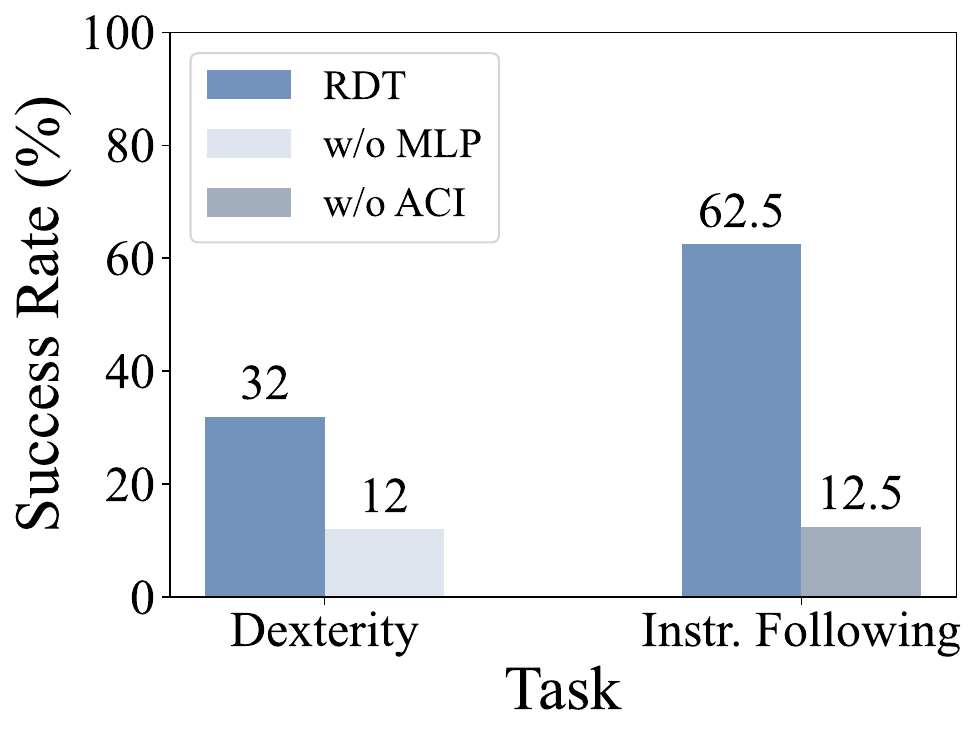}
     % \vspace{-0.5em}
    \subcaption{Task w/o MLP or ACI}
    \vspace{-0.5ex}
    \label{fig:necessity:b}
\end{minipage}
% \vspace{-2ex}
% \includegraphics[width=1.0\linewidth]{loss.pdf}%\vspace{-1ex}
\captionsetup{font=small}
\caption{{\textbf{(a)}  Unstable loss curve during training without QKNorm \& RMSNorm. \textbf{(b)} Success rates of RDT (w/o MLP Decoder or w/o ACI) in tasks of \textit{Robot Dog} (walk straight sub-task) and \textit{Pour Water-L-1/3} (correct amount sub-task). See Fig.~\ref{fig:task} for task definitions. All the models are without pre-training in this experiment due to resource constraints.}}
\label{fig:necessity}
\vspace{-0.5cm}
\end{wrapfigure}

% {\bf }
\paragraph{Network Structure of $f_{\vth}$.}
We choose Transformer as the scalable backbone network~\citep{bao2023all,peebles2023scalable} and make the following three key modifications from Diffusion Transfomer (DiT) by considering the characteristics of our robotic problem: 
\begin{itemize}[leftmargin=1.0em]
\vspace{-0.5em}
\setlength\itemsep{0em}
    \item \textbf{QKNorm \& RMSNorm.} The \textit{unstable numerical range} of the inputting robotic physical quantities can lead to problems such as gradient instability and numerical overflow, especially when training large foundation models. To solve this problem, we add QKNorm~\citep{henry2020query} to avoid numerical instability when calculating attention. Besides, we also note that our problem can be considered as a time series forecasting task, and the centering operation in the original DiTs' LayerNorm could cause \emph{token shift} and \emph{attention shift}, thus destroying the symmetry of the time series~\citep{huang2024unitnorm}. Therefore, we replace LayerNorm with RMSNorm~\citep{zhang2019root} without a centering operation. Fig.~\ref{fig:necessity:a} shows that large-scale pre-training tends to be very unstable or even explode without this modification.
    
    \item \textbf{MLP Decoder.} To improve the approximation capability for \textit{nonlinear} robot actions, we replace the final linear decoder with a nonlinear MLP decoder as a projection from the latent space back to the physical space. As empirically shown in Fig.~\ref{fig:necessity:b}, without this design, RDT cannot effectively capture nonlinear dynamics and thus loses the ability to accomplish dexterous tasks that require delicate operations.

    \item \textbf{Alternating Condition Injection (ACI).} In our model, image and language inputs serve as conditions, which are high-dimensional and variable in length, contrasting with the class label conditions in \modify{traditional DiTs~\citep{peebles2023scalable}}. These informative conditions are challenging to compress into a single token, making the original adaptive layer norm approach unsuitable. Therefore, we employ cross-attention to accommodate conditions of varying lengths avoiding the information loss in further compression. Besides, we further analyze that, given that image tokens are usually much more than text tokens, simultaneous injection of both modalities tends to overshadow text-related information, thus impairing the capability of the instruction following (see Fig.~\ref{fig:necessity:b} for quantitative results). To mitigate this issue, we strategically alternate between injecting image and text tokens in successive layers' cross-attention rather than injecting both in every layer.

\vspace{-0.5em}
\end{itemize}

\subsection{Data}\label{sec:model:data}
\paragraph{Training on Heterogeneous Multi-Robot Data.}
To enable training on heterogeneous multi-robot data, \modify{we need a unified action space shared among various robots to provide a unified format for multi-robot actions.} The mapping from the original action space of a robot to the unified action space should be physically interpretable, \modify{and each dimension of the space should have a clear physical meaning.} This can encourage the model to learn shared physical laws from different robot data, thereby improving the efficiency of learning from data of different robots~\citep{shah2023gnm}. 

The design of the space consists of two steps. Firstly, for each robot, we can use a single space to accommodate both its proprioception $\vz_{t}$ and action $\va_t$. This is because $\va_t$ is usually a subset of the desired $\vz_{t+1}$~\citep{de2012theory,kouvaritakis2016model}, and thus the space of $\vz_{t}$ naturally contains the space of $\va_t$. Secondly, we design a unified space that encompasses all the main physical quantities of most robots with gripper arms. As illustrated in the left side of Fig.~\ref{fig:framework}, we embed the action space of a robot into this unified space by filling each element of the original action vector into the corresponding position of the unified action space vector according to its physical meaning, with the remaining positions being padded. The specific definition of the space is given in App.~\ref{unified_action_space}. 

With this unified space, we are able to pre-train RDT on data from almost all modern robots with gripper arms, and greatly expand the data scale towards the requirement for a foundation model. Specifically, our collection of pre-training datasets includes $46$ datasets of various robots, with a total size of $1$M+ trajectories and $21$TB. More details and preprocessing are deferred to App.~\ref{pretrain_dataset_detail}.

\paragraph{Collecting a Comprehensive Multi-Task Bimanual Dataset.}
Though having been pre-trained on large-scale datasets, RDT could still need help to zero-shot generalize to the target dual-arm robot due to the embodiment gap. To bridge the gap, we need to collect a multi-task bimanual dataset on the target robot for fine-tuning. Recent advances in large language models~\citep{ziegler2019fine,brown2020language,touvron2023llama}  have shown that high-quality fine-tuning datasets are crucial for model performance. We ensure the high quality of our dataset from three aspects: \textbf{(1)} Regarding quantity, we have collected $6$K+ trajectories, making our dataset one of the largest bimanual datasets nowadays; \textbf{(2)} Regarding comprehensiveness, we consider $300$+ challenging tasks, covering most manipulation task types, from pick-and-place to plugging cables, even including writing math equations; \textbf{(3)} Regarding diversity, we prepare $100$+ objects with rigid and non-rigid bodies of various sizes and textures and $15$+ different rooms with different lighting conditions. Besides, we further utilize GPT-4-Turbo~\citep{achiam2023gpt} to rewrite human-annotated instructions to increase text diversity. For more information, we refer to Fig.~\ref{fig:ft_dataset} and App.~\ref{app:ft_dataset}.

\section{Experiments}
\label{sec:experiments}
We aim to answer the following questions through real-robot experiments: $\mathcal{Q}$\textbf{1}: Can RDT zero-shot generalize to unseen objects and scenes? $\mathcal{Q}$\textbf{2}: How effective is RDT's zero-shot instruction-following capability for unseen modalities? $\mathcal{Q}$\textbf{3}: Can RDT facilitate few-shot learning for previously unseen skills? $\mathcal{Q}$\textbf{4}: Is RDT capable of completing tasks that require delicate operations? and $\mathcal{Q}$\textbf{5}: Are large model sizes, extensive data, and diffusion modeling helpful for RDT's performance? 
\iffalse
\begin{enumerate}[leftmargin=2.5em,label=${\mathcal{Q}}$\textbf{\arabic*:}]
\vspace{-0.5em}
\setlength\itemsep{0em}
    \item Can RDT zero-shot generalize to unseen objects and scenes?
    \item How effective is RDT's zero-shot instruction-following capability for unseen modalities?
    \item Can RDT facilitate few-shot learning for previously unseen skills?
    \item Is RDT capable of completing tasks that require delicate operations?
    \item Are large model sizes, extensive data, and diffusion modeling helpful for RDT's performance?
    
    % large-scale pre-training help improve the performance and generalizability of RDT?
\end{enumerate}
\fi
\subsection{Experiment Setups}\label{sec:exp:setup}

\begin{figure}[t]
    \centering
    \includegraphics[width=\textwidth]{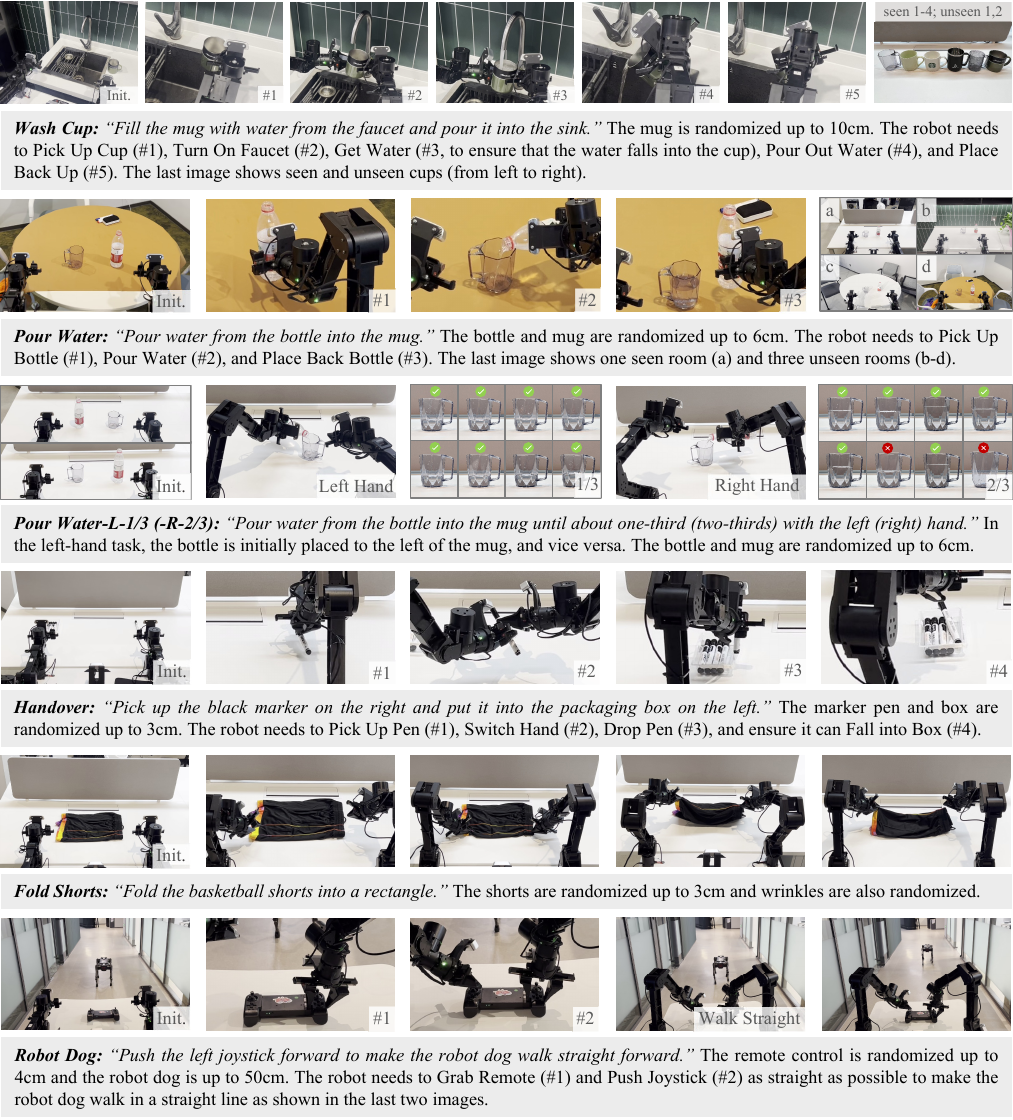}
    % \fbox{\rule{0pt}{2.0in} \rule{0.95\linewidth}{0pt}}
    \vspace{-1em}
    \caption{\textbf{Task definitions and visualizations.} For $7$ challenging tasks, we describe their language instruction, randomization, and definitions of each sub-task. For \textit{Pour Water-L-1/3} and \textit{Pour Water-R-2/3}, we show the resulting water levels in two images.}
    \label{fig:task}\vspace{-2ex}
\end{figure}

\begin{table}[t]
\caption{\textbf{Dimensions when designing tasks.} For \textit{Pour Water-L-1/3} and \textit{Pour Water-R-2/3}, only the water levels of \textit{little}, \textit{half} (i.e., 1/2), and \textit{full} are seen in training instructions. For \textit{Handover} and \textit{Fold Shorts}, the dataset only contains $5$ demos and $1$ demo of the skill, respectively. For \textit{Robot Dog}, it requires delicate operations, as a slight angle when pushing joysticks can make the robot dog deviate.}
\vspace{-1em}
\label{tbl:taskdim}
\renewcommand{\arraystretch}{1.2}
\begin{center}
\resizebox{\linewidth}{!}{
\begin{tabular}{lll}
\toprule
\multicolumn{1}{c}{\textbf{TASK NAME}} & \multicolumn{1}{c}{\textbf{DIMENSION}}             & \multicolumn{1}{c}{\textbf{EXPLANATION}}                                                   \\ \hline
Wash Cup                      & Unseen Object ($\mathcal{Q}$\textbf{1})        & To wash one seen and two unseen cups with the faucet                              \\
Pour Water                    & Unseen Scene ($\mathcal{Q}$\textbf{1})         & To pour water into the cup in three unseen rooms                                  \\
Pour Water-L-1/3              & Instruction Following ($\mathcal{Q}$\textbf{2}) & To pour water into the cup with the \textbf{left hand} until \textbf{one-third }full                \\
Pour Water-R-2/3              & Instruction Following ($\mathcal{Q}$\textbf{2}) & To pour water into the cup with the \textbf{right hand} until \textbf{two-thirds} full              \\
Handover                      & 5-Shot Learning ($\mathcal{Q}$\textbf{3})       & To move the marker to the box, where handover is needed due to far distance          \\
Fold Shorts                   & 1-Shot Learning ($\mathcal{Q}$\textbf{3})       & To fold the shorts in half horizontally                                           \\
Robot Dog                     & Dexterity ($\mathcal{Q}$\textbf{4})                                       & To push the joystick straight to control the robot dog to walk in a straight line \\ \bottomrule
\end{tabular}
}
\end{center}
\vspace{-2ex}
\end{table}

\paragraph{Tasks.}
We select $7$ challenging tasks to evaluate the generalizability and capabilities of RDT from different dimensions, including complex scenarios that the model may encounter in real-world tasks, such as various unseen elements and dexterous manipulation. An illustration of the dimension of each task is given in Table~\ref{tbl:taskdim} while detailed definitions and visualizations are provided in Fig.~\ref{fig:task}.

\paragraph{Data.}
We use the pre-training and fine-tuning datasets in Sec.~\ref{sec:model:data}. We now list the number of demos related to each task in our fine-tuning dataset. \textit{Wash Cup}: $133$ demos for seen cups combined and $0$ demos for unseen cups; \textit{Pour Water}: $350$ demos for seen rooms combined and $0$ demos for unseen rooms; \textit{Pour Water-L-1/3} \& \textit{Pour Water-R-2/3}: $18$ demos for the water level of little, $19$ demos for half, and $19$ demos for full; \textit{Handover}: $5$ demos; \textit{Fold Shorts}: $1$ demo; \textit{Robot Dog}: $68$ demos.

% \paragraph{Scaling Up.}
% As a foundation model, RDT is expected to learn as much knowledge as possible from the data. To this end, we perform an all-around scaling up, including model size, computational resources, and training iterations. We scale the size of RDT up to $1.2$B parameters, establishing it as the currently largest open-source diffusion-based robotic foundation model. The model is trained on $48$ H100 80GB GPUs for a month, giving a total of $1$M training iteration steps. It takes three days to fine-tune this model using the same GPUs on our fine-tuning dataset for $130$K steps. We defer further training details to App.~\ref{app:train}, including detailed design choices and anti-overfitting techniques.

% To speed up the inference of RDT, we have adopted a recent advance in sampling acceleration techniques of diffusion models, DPM-Solver++~\citep{lu2022dpm}. This technique can reduce the diffusion steps required to sample an action chunk from $100$ steps to $5$ steps. In experiments, we have deployed RDT on the target dual-arm robot equipped with an RTX 4090 24GB GPU (see App.~\ref{hardware_detail} for hardware configurations). It can achieve an action chunk inference frequency of $6$ Hz (action chunks per second) and an average action inference frequency of $381$ Hz (actions per second). This speed is comparable to human operators and is sufficient for most bimanual tasks.

{\bf Model Training and Inference.}
We scale the size of RDT up to $1.2$B parameters, establishing it as the currently largest diffusion-based robotic foundation model. The model is pre-trained on $48$ H100 80GB GPUs for a month, giving a total of $1$M training iteration steps. It takes three days to fine-tune this model using the same GPUs for $130$K steps. We defer further details to App.~\ref{app:train}, including the running platform, design choices, and data augmentation techniques. For real-time inference, we adopt DPM-Solver++~\citep{lu2022dpm}, a recent sampling accelerator of diffusion models. It can reduce the diffusion steps required to sample an action chunk from $100$ steps to $5$ steps, achieving an action chunk inference frequency of $6$ Hz (action chunks per second) and an average action inference frequency of $381$ Hz (actions per second) on the target robot's onboard RTX 4090 24GB GPU.

{\bf Baselines.}
To comprehensively evaluate RDT, we consider the most advanced baselines in robotic foundation models and bimanual manipulation, including Action Chunking with Transformers (ACT)~\citep{zhao2023learning}, OpenVLA~\citep{kim2024openvla}, and Octo~\citep{team2023octo}. ACT is a state-of-the-art method in bimanual manipulation, which uses VAE to model the action distribution. OpenVLA is the largest open-source foundation model (7B), employing the discretization modeling. Octo is a diffusion-based foundation model, and its largest version has only $93$M parameters.

% a representative model for bimanual manipulation and is also the state-of-the-art method evaluated on the ALOHA robot. Rather than diffusion, it uses VAE to model the action distribution. OpenVLA is the largest open-source vision-language-action model (7B), which discretizes the action space and models the action distribution with categorical probabilities. 

% Due to the lack of unified action space, it has been pre-trained on only unimanual data, giving a total of $970$k episodes.

\paragraph{Metric and Hardware.}
We employ the success rate as our main metric, which is calculated by dividing successful trials by total trials. \textit{Wash Cup} is tested with $8$ trials for each cup (one seen cup, two unseen cups, $24$ trials in total). \textit{Pour Water} is tested with $8$ trials for each room (three unseen rooms, $24$ trials in total). \textit{Pour Water-L-1/3} and \textit{Pour Water-R-2/3} are tested with $8$ trials each. \textit{Handover}, \textit{Fold Shorts}, and \textit{Robot Dog} are tested with $25$ trials each. All the tests are performed on the ALOHA dual-arm robot (see App.~\ref{hardware_detail} for hardware configurations). Experimental details, such as the implementation and hyper-parameters, are elaborated in App.~\ref{app:exp}.

\begin{wraptable}{r}{0.5\linewidth}
% \vspace{-0.3em}
\centering
\captionsetup{font=small}
\caption{\textbf{Ablation study results.} Here are the success rates ($\%$) of the original RDT and its three variants in tasks of \textit{Wash Cup} (unseen cup 2, total success rate), \textit{Pour Water} (unseen room 3, total success rate), and \textit{Pour Water-L-1/3} (correct amount sub-task). All the models except \textit{RDT (scratch)} are pre-trained before fine-tuning.}
\label{tbl:ablation}
\renewcommand{\arraystretch}{1.2}
\begin{center}
\resizebox{\linewidth}{!}{
\begin{tabular}{lccc}
\hline
\multicolumn{1}{c}{\textbf{\begin{tabular}[c]{@{}c@{}}VARIANT\\ NAME\end{tabular}}} & \textbf{\begin{tabular}[c]{@{}c@{}}UNSEEN\\ OBJECT\end{tabular}} & \textbf{\begin{tabular}[c]{@{}c@{}}UNSEEN\\ SCENE\end{tabular}} & \multicolumn{1}{c}{\textbf{\begin{tabular}[c]{@{}c@{}}INSTRUCTION\\ FOLLOWING\end{tabular}}} \\ \hline
RDT (regress)                                                                       & 12.5                                                             & 50                                                              & 12.5                                                                                         \\
RDT (small)                                                                         & 37.5                                                             & \textbf{62.5}                                                            & 25                                                                                           \\
RDT (scratch)                                                                       & 0                                                                & 25                                                              & 62.5                                                                                         \\
RDT (\textbf{ours})                                                                          & \textbf{50}                                                      & \textbf{62.5}                                                   & \textbf{100}                                                                                 \\ \hline
\end{tabular}
}
\end{center}
\end{wraptable}

\paragraph{Ablation Study.}
Answering $\mathcal{Q}$\textbf{5}, we have conducted ablation studies on the model size, pre-training, and the modeling method to understand their importance. We consider the variants of: \textit{RDT (ours):} the original RDT. \textit{RDT (regress):} RDT without diffusion modeling. It models the deterministic mapping $(\ell, \vo_t) \mapsto \va_{t}$. \textit{RDT (small):} RDT without large parameters. It has only $166$M parameters. \textit{RDT (scratch):} RDT without pre-training. It is trained from scratch during fine-tuning. In Table~\ref{tbl:ablation}, we evaluate these variants in terms of three dimensions of generalizability. Table~\ref{tbl:baseline} provides a comparison of different variants of RDT as well as baselines.

\begin{table}[t]
\def\Thickhline{\noalign{\hrule height1pt}}
% thick line, large table head, smaller spacing in "Pick Up // Cup"

\caption{\textbf{Quantitative results.} We report success rates ($\%$) of ACT, OpenVLA, RDT (from scratch, no pre-trained), and RDT (ours, pre-trained) for $7$ tasks. Sub-columns in each sub-task cell represent different elements (objects, instructions, scenes). ACT is not language-conditioned and thus unavailable for instruction following. RDT (\textbf{ours}) consistently outperforms others.}
\vspace{-1em}
\label{tbl:result}
\renewcommand{\arraystretch}{1.2}
\begin{center}
\resizebox{\linewidth}{!}{
\begin{tabular}{lcccccccccccccccccc}
\Thickhline
                     & \multicolumn{18}{c}{Wash Cup: seen cup 1 $|$ unseen cup 1 $|$ unseen cup 2 (\textbf{Unseen Object})}                                                                                                                                                                                                                                                                                                                                                                                                                                                                                                                                                                                                 \\ \hline
                     & \multicolumn{3}{c|}{\renewcommand{\arraystretch}{1.1}\begin{tabular}[c]{@{}c@{}}\footnotesize{Pick Up} \footnotesize{Cup}\end{tabular}}                                                & \multicolumn{3}{c|}{\renewcommand{\arraystretch}{1.1}\begin{tabular}[c]{@{}c@{}}\footnotesize{Turn On} \footnotesize{Faucet}\end{tabular}}                      & \multicolumn{3}{c|}{\renewcommand{\arraystretch}{1.1}\begin{tabular}[c]{@{}c@{}}\footnotesize{Get} \footnotesize{Water}\end{tabular}}                                             & \multicolumn{3}{c|}{\renewcommand{\arraystretch}{1.1}\begin{tabular}[c]{@{}c@{}}\footnotesize{Pour Out} \footnotesize{Water}\end{tabular}}                 & \multicolumn{3}{c|}{\renewcommand{\arraystretch}{1.1}\begin{tabular}[c]{@{}c@{}}\footnotesize{Place Back} \footnotesize{Cup}\end{tabular}}                                    & \multicolumn{3}{c}{\footnotesize{Total}}                                                                        \\
ACT                  & 50                                  & 12.5                                & \multicolumn{1}{c|}{37.5}                     & 0                         & 0                        & \multicolumn{1}{c|}{0}                      & 0                                    & 0                                    & \multicolumn{1}{c|}{0}                 & 0                 & 0                        & \multicolumn{1}{c|}{0}                         & 37.5                                 & 0                                    & \multicolumn{1}{c|}{0}             & 0                  & 0                          & 0                                              \\
OpenVLA              & 0                                   & 0                                   & \multicolumn{1}{c|}{0}                        & 0                         & 0                        & \multicolumn{1}{c|}{0}                      & 0                                    & 0                                    & \multicolumn{1}{c|}{0}                 & 0                 & 0                        & \multicolumn{1}{c|}{0}                         & 0                                    & 0                                    & \multicolumn{1}{c|}{0}             & 0                  & 0                          & 0                                              \\
Octo              & 0                                   & 0                                   & \multicolumn{1}{c|}{0}                        & 0                         & 0                        & \multicolumn{1}{c|}{0}                      & 0                                    & 0                                    & \multicolumn{1}{c|}{0}                 & 0                 & 0                        & \multicolumn{1}{c|}{0}                         & 0                                    & 0                                    & \multicolumn{1}{c|}{0}             & 0                  & 0                          & 0                                              \\
RDT (scratch)        & 37.5                                & 12.5                                & \multicolumn{1}{c|}{0}                        & 0                         & 12.5                     & \multicolumn{1}{c|}{12.5}                   & 0                                    & 0                                    & \multicolumn{1}{c|}{0}                 & 37.5              & 12.5                     & \multicolumn{1}{c|}{0}                         & 25                                   & 0                                    & \multicolumn{1}{c|}{0}             & 0                  & 0                          & 0                                              \\
RDT (\textbf{ours})           & 87.5                                & 87.5                                & \multicolumn{1}{c|}{50}                       & 62.5                      & 75                       & \multicolumn{1}{c|}{50}                     & 50                                   & 75                                   & \multicolumn{1}{c|}{50}                & 87.5              & 75                       & \multicolumn{1}{c|}{50}                        & 87.5                                 & 62.5                                 & \multicolumn{1}{c|}{50}            & \textbf{50}        & \textbf{75}                & \textbf{50}                                    \\ \hline
                     & \multicolumn{18}{c}{Pour Water-L-1/3 $|$ Pour Water-R-2/3 (\textbf{Instruction Following})}                                                                                                                                                                                                                                                                                                                                                                                                                                                                                                                                                                                          \\ \hline
                     & \multicolumn{3}{c|}{\renewcommand{\arraystretch}{1.1}\begin{tabular}[c]{@{}c@{}}\footnotesize{Pick Up} \footnotesize{Bottle}\end{tabular}}                                             & \multicolumn{3}{c|}{\renewcommand{\arraystretch}{1.1}\begin{tabular}[c]{@{}c@{}}\footnotesize{Pour} \footnotesize{Water}\end{tabular}}                          & \multicolumn{3}{c|}{\renewcommand{\arraystretch}{1.1}\begin{tabular}[c]{@{}c@{}}\footnotesize{Place Back} \footnotesize{Bottle}\end{tabular}}                                     & \multicolumn{3}{c|}{\footnotesize{Total}}                                                                    & \multicolumn{3}{c|}{\renewcommand{\arraystretch}{1.1}\begin{tabular}[c]{@{}c@{}}\footnotesize{Correct} \footnotesize{Hand}\end{tabular}}                                      & \multicolumn{3}{c}{\renewcommand{\arraystretch}{1.1}\begin{tabular}[c]{@{}c@{}}\footnotesize{Correct} \footnotesize{Amount}\end{tabular}}                     \\
OpenVLA              & 50                                  &                                     & \multicolumn{1}{c|}{0}                        & 0                         &                          & \multicolumn{1}{c|}{0}                      & 0                                    &                                      & \multicolumn{1}{c|}{0}                 & 0                 &                          & \multicolumn{1}{c|}{0}                         & 50                                   &                                      & \multicolumn{1}{c|}{0}             & 0                  &                            & \multicolumn{1}{c}{0}                          \\
Octo              &  0                                 &                                     & \multicolumn{1}{c|}{0}                        & 0                         &                          & \multicolumn{1}{c|}{0}                      & 0                                    &                                      & \multicolumn{1}{c|}{0}                 & 0                 &                          & \multicolumn{1}{c|}{0}                         & 0                                   &                                      & \multicolumn{1}{c|}{0}             & 0                  &                            & \multicolumn{1}{c}{0}                          \\
RDT (scratch)        & 100                                 &                                     & \multicolumn{1}{c|}{75}                       & 75                        &                          & \multicolumn{1}{c|}{25}                     & 62.5                                 &                                      & \multicolumn{1}{c|}{25}                & 62.5              &                          & \multicolumn{1}{c|}{25}                        & 100                                  &                                      & \multicolumn{1}{c|}{75}            & 62.5               &                            & \multicolumn{1}{c}{12.5}                       \\
RDT (\textbf{ours})           & 100                                 &                                     & \multicolumn{1}{c|}{87.5}                     & 100                       &                          & \multicolumn{1}{c|}{87.5}                   & 100                                  &                                      & \multicolumn{1}{c|}{87.5}              & \textbf{100}      &                          & \multicolumn{1}{c|}{\textbf{87.5}}             & \textbf{100}                         &                                      & \multicolumn{1}{c|}{\textbf{87.5}} & \textbf{100}      &                            & \multicolumn{1}{c}{\textbf{75}}                \\ \hline
\multicolumn{1}{c}{} & \multicolumn{12}{c}{Pour Water: unseen room 1 $|$ unseen room 2 $|$ unseen room 3 (\textbf{Unseen Scene})}                                                                                                                                                                                                                                                                                                                                                                    & \multicolumn{1}{c}{}                 & \multicolumn{1}{c}{}                 & \multicolumn{1}{c}{}               & \multicolumn{3}{c}{Fold Shorts (\textbf{1-Shot})}                                                         \\ \hline
                     & \multicolumn{3}{c|}{\renewcommand{\arraystretch}{1.1}\begin{tabular}[c]{@{}c@{}}\footnotesize{Pick Up} \footnotesize{Bottle}\end{tabular}}                                             & \multicolumn{3}{c|}{\renewcommand{\arraystretch}{1.1}\begin{tabular}[c]{@{}c@{}}\footnotesize{Pour} \footnotesize{Water}\end{tabular}}                          & \multicolumn{3}{c|}{\renewcommand{\arraystretch}{1.1}\begin{tabular}[c]{@{}c@{}}\footnotesize{Place Back} \footnotesize{Bottle}\end{tabular}}                                     & \multicolumn{3}{c|}{\footnotesize{Total}}                                                                    & \multicolumn{3}{c|}{-}                                                                                           & \multicolumn{3}{c}{\footnotesize{Total}}                                                                        \\
ACT                  & 25                                  & 87.5                                & \multicolumn{1}{c|}{25}                       & 0                         & 50                       & \multicolumn{1}{c|}{12.5}                   & 0                                    & 37.5                                 & \multicolumn{1}{c|}{12.5}              & 0                 & 37.5                     & \multicolumn{1}{c|}{12.5}                      & \multicolumn{1}{c}{}                 & -                                    & \multicolumn{1}{c|}{}              & \multicolumn{3}{c}{0}                                                                            \\
OpenVLA              & 0                                   & 0                                   & \multicolumn{1}{c|}{0}                        & 0                         & 0                        & \multicolumn{1}{c|}{0}                      & 0                                    & 0                                    & \multicolumn{1}{c|}{0}                 & 0                 & 0                        & \multicolumn{1}{c|}{0}                         & \multicolumn{1}{c}{}                 & -                                    & \multicolumn{1}{c|}{}              & \multicolumn{3}{c}{0}                                                                            \\
Octo              & 50                                   & 0                                   & \multicolumn{1}{c|}{12.5}                        & 12.5                         & 0                        & \multicolumn{1}{c|}{0}                      & 12.5                                    & 0                                    & \multicolumn{1}{c|}{0}                 & 12.5                 & 0                        & \multicolumn{1}{c|}{0}                         & \multicolumn{1}{c}{}                 & -                                    & \multicolumn{1}{c|}{}              & \multicolumn{3}{c}{4}                                                                            \\
RDT (scratch)        & 62.5                                & 100                                 & \multicolumn{1}{c|}{62.5}                     & 25                        & 87.5                     & \multicolumn{1}{c|}{37.5}                   & 25                                   & 75                                   & \multicolumn{1}{c|}{25}                & 25                & 75                       & \multicolumn{1}{c|}{25}                        & \multicolumn{1}{c}{}                 & -                                    & \multicolumn{1}{c|}{}              & \multicolumn{3}{c}{40}                                                                           \\
RDT (\textbf{ours})           & 62.5                                & 100                                 & \multicolumn{1}{c|}{62.5}                     & 62.5                      & 100                      & \multicolumn{1}{c|}{62.5}                   & 62.5                                 & 100                                  & \multicolumn{1}{c|}{62.5}              & \textbf{62.5}     & \textbf{100}             & \multicolumn{1}{c|}{\textbf{62.5}}             & \multicolumn{1}{c}{}                 & -                                    & \multicolumn{1}{c|}{}              & \multicolumn{3}{c}{\textbf{68}}                                                                  \\ \hline
                     & \multicolumn{10}{c}{Handover (\textbf{5-Shot})}                                                                                                                                                                                                                                                                                                                                    & \multicolumn{8}{c}{Robot Dog (\textbf{Dexterity})}                                                                                                                                                                                                                                                        \\ \hline
                     & \multicolumn{2}{c}{\renewcommand{\arraystretch}{1.1}\begin{tabular}[c]{@{}c@{}}\footnotesize{Pick Up}\\ \footnotesize{Pen}\end{tabular}} & \multicolumn{2}{c}{\renewcommand{\arraystretch}{1.1}\begin{tabular}[c]{@{}c@{}}\footnotesize{Switch}\\ \footnotesize{Hand}\end{tabular}} & \multicolumn{2}{c}{\renewcommand{\arraystretch}{1.1}\begin{tabular}[c]{@{}c@{}}\footnotesize{Drop}\\ \footnotesize{Pen}\end{tabular}} & \multicolumn{2}{c}{\renewcommand{\arraystretch}{1.1}\begin{tabular}[c]{@{}c@{}}\footnotesize{Fall into}\\ \footnotesize{Box}\end{tabular}} & \multicolumn{2}{c|}{\footnotesize{Total}}                                 & \multicolumn{2}{c}{\renewcommand{\arraystretch}{1.1}\begin{tabular}[c]{@{}c@{}}\footnotesize{Grab}\\ \footnotesize{Remote}\end{tabular}} & \multicolumn{2}{c}{\renewcommand{\arraystretch}{1.1}\begin{tabular}[c]{@{}c@{}}\footnotesize{Push}\\ \footnotesize{Joystick}\end{tabular}} & \multicolumn{2}{c}{\footnotesize{Total}}                               & \multicolumn{2}{c}{\renewcommand{\arraystretch}{1.1}\begin{tabular}[c]{@{}c@{}}\footnotesize{Walk}\\ \footnotesize{Straight}\end{tabular}} \\
ACT                  & \multicolumn{2}{c}{44}                                                    & \multicolumn{2}{c}{0}                                                     & \multicolumn{2}{c}{0}                                                  & \multicolumn{2}{c}{0}                                                       & \multicolumn{2}{c|}{0}                                     & \multicolumn{2}{c}{88}                                                    & \multicolumn{2}{c}{32}                                                      & \multicolumn{2}{c}{32}                                  & \multicolumn{2}{c}{32}                                                      \\
OpenVLA              & \multicolumn{2}{c}{0}                                                     & \multicolumn{2}{c}{0}                                                     & \multicolumn{2}{c}{0}                                                  & \multicolumn{2}{c}{0}                                                       & \multicolumn{2}{c|}{0}                                     & \multicolumn{2}{c}{84}                                                    & \multicolumn{2}{c}{0}                                                       & \multicolumn{2}{c}{0}                                   & \multicolumn{2}{c}{0}                                                       \\
Octo              & \multicolumn{2}{c}{12}                                                     & \multicolumn{2}{c}{0}                                                     & \multicolumn{2}{c}{0}                                                  & \multicolumn{2}{c}{0}                                                       & \multicolumn{2}{c|}{0}                                     & \multicolumn{2}{c}{100}                                                    & \multicolumn{2}{c}{4}                                                       & \multicolumn{2}{c}{4}                                   & \multicolumn{2}{c}{0}                                                       \\
RDT (scratch)        & \multicolumn{2}{c}{88}                                                    & \multicolumn{2}{c}{32}                                                    & \multicolumn{2}{c}{24}                                                 & \multicolumn{2}{c}{16}                                                      & \multicolumn{2}{c|}{16}                                    & \multicolumn{2}{c}{100}                                                   & \multicolumn{2}{c}{64}                                                      & \multicolumn{2}{c}{64}                                  & \multicolumn{2}{c}{32}                                                      \\
RDT (\textbf{ours})           & \multicolumn{2}{c}{100}                                                   & \multicolumn{2}{c}{56}                                                    & \multicolumn{2}{c}{56}                                                 & \multicolumn{2}{c}{40}                                                      & \multicolumn{2}{c|}{\textbf{40}}                           & \multicolumn{2}{c}{100}                                                   & \multicolumn{2}{c}{76}                                                      & \multicolumn{2}{c}{\textbf{76}}                         & \multicolumn{2}{c}{\textbf{48}}                                             \\ \Thickhline
\end{tabular}
}
\end{center}
\vspace{-2ex}
\end{table}

\subsection{Results Analysis}
From the results in Table~\ref{tbl:result}, we can see that RDT consistently outperforms other baselines. This is because RDT employs diffusion with a powerful network architecture to model the distribution of multi-modal actions accurately, while discretization and VAE lack accuracy and expressiveness, respectively. Besides, the large number of parameters after large-scale pre-training provides a lot of prior knowledge, which significantly improves the generalizability. Here is a detailed analysis:
\begin{itemize}[leftmargin=1.0em]
\vspace{-0.5em}
\setlength\itemsep{0em}
\item \textbf{$\mathcal{Q}$1 \& $\mathcal{Q}$2:} RDT can zero-shot generalize to unseen objects, scenes, and modalities. In \textit{Wash Cup} and \textit{Pour Water}, RDT can still achieve a high success rate on unseen scenarios, and its performance is not much different from that on seen ones. In contrast, the other baselines cannot even complete the entire task. In \textit{Pour Water-L-1/3} and \textit{Pour Water-R-2/3}, from the third row of Fig.~\ref{fig:task} or Fig.~\ref{fig:water} (zoomed-in version), we can find that RDT understands precisely which hand to manipulate and how much water to pour and closely follows the instruction through its actions, even though it has never seen words like ``one-third" or ``two-thirds". It is precisely because of large-scale pre-training that RDT has seen a large number of diverse objects, scenes, and instructions, leading to such strong zero-shot generalization.
\item \textbf{$\mathcal{Q}$3:} RDT can learn new skills using only a few shots. In \textit{Handover} and \textit{Fold Shorts}, RDT has learned new and complex skills of handover and folding through few-shot learning, whose action patterns are very different from known skills, while the success rate of others is almost zero. Such improvement is also due to large-scale pre-training. Few-shot learning can help RDT quickly adapt to new working environments, which is of great significance for practical applications.
\item \textbf{$\mathcal{Q}$4:} RDT can handle dexterous tasks. In \textit{Robot Dog}, RDT accurately controls the angle when pushing the joystick, while others have caused the robot dog to deviate. This is because diffusion, with our powerful network architecture, can model the distribution of multi-modal and nonlinear actions so that the action precision can meet the requirements of dexterous tasks. We also note that the joystick and the remote control are both black, making the joystick not visually apparent. It probably makes ACT prone to failure. In contrast, large-scale pre-training has made RDT learn a better vision-language representation of the joystick concept, improving the recognition capability.
\item \textbf{$\mathcal{Q}$5:} Large model size, extensive data, and diffusion are all essential factors for our excellence. In Table~\ref{tbl:ablation}, there is a serious performance drop without any of these factors, demonstrating the necessity of our contributions. In particular, \textit{RDT (scratch)} performs poorly on unseen objects and scenes, indicating that the knowledge from pre-training is critical for generalization.
\end{itemize}

\section{Conclusion}
\label{sec:conclusion}
In this paper, we tackled the challenges of data scarcity and increased manipulation complexity in generalizable bimanual manipulation by developing the Robotics Diffusion Transformer (RDT), a diffusion-based foundation model for language-conditioned visuomotor imitation learning. Our model was pre-trained on an extensive multi-robot dataset and fine-tuned on a self-collected bimanual dataset. We further introduce a Physically Interpretable Unified Action Space to unify action representations across different robots, enhancing robustness and transferability. Outperforming existing methods, RDT not only demonstrates significant improvements in dexterous bimanual capability and instruction following but also achieves remarkable performance in few-shot learning and zero-shot generalization to unseen objects and scenes.

% generalization capabilities.

% , setting new performance benchmarks in dexterity, coordination, and adaptability

\clearpage

% \subsection*{Author Contributions}

\subsection*{Acknowledgments}
This work was supported by NSFC Projects (Nos. 92248303, 92370124, 62350080, 62276149,
62061136001, 92270001), BNRist (BNR2022RC01006), Tsinghua Institute for Guo Qiang, and the
High Performance Computing Center, Tsinghua University. J. Zhu was also supported by the XPlorer
Prize.
\subsection*{Ethics Statement}

All the data used in this research comes from open-source and well-documented datasets, and we strictly follow all applicable licensing and usage guidelines. Our finetuning dataset is collected by the authors of this paper along with some volunteers.

While RDT is a model trained for scalable, language-conditioned visuomotor policy learning and tested on the ALOHA dual-arm robot, we emphasize that any harmful use of our model is neither intended nor encouraged, and we encourage responsible deployment on real-world robots.

\subsection*{Reproducibility Statement}

% All our code, model weights, and datasets are fully open-source for reproducibility at the link. 

To reproduce our pre-training and fine-tuning processes, we have provided the code in the \href{https://github.com/thu-ml/RoboticsDiffusionTransformer}{repository}. We also include instructions for downloading the dataset, how to use the training code, and a guide for deploying on a real machine in the README file. We have fully open-sourced all our code, model weights, and fine-tuning datasets. We refer to the \href{https://rdt-robotics.github.io/rdt-robotics/}{project page} for more information.

Please refer to App.~\ref{pretrain_dataset_detail} for pre-training dataset details, App.~\ref{app:ft_dataset} for fine-tuning dataset details, App.~\ref{app:train} for RDT training details, App.~\ref{hardware_detail} for hardware details, and App.~\ref{app:exp} for experimental details and implementation of baselines.

\bibliography{iclr2024_conference}

\begin{thebibliography}{113}
\providecommand{\natexlab}[1]{#1}
\providecommand{\url}[1]{\texttt{#1}}
\expandafter\ifx\csname urlstyle\endcsname\relax
  \providecommand{\doi}[1]{doi: #1}\else
  \providecommand{\doi}{doi: \begingroup \urlstyle{rm}\Url}\fi

\bibitem[TFD()]{TFDS}
{TensorFlow Datasets}, a collection of ready-to-use datasets.
\newblock \url{https://tensorflow.google.cn/datasets}.

\bibitem[Abadi et~al.(2015)Abadi, Agarwal, Barham, Brevdo, Chen, Citro, Corrado, Davis, Dean, Devin, Ghemawat, Goodfellow, Harp, Irving, Isard, Jia, Jozefowicz, Kaiser, Kudlur, Levenberg, Man\'{e}, Monga, Moore, Murray, Olah, Schuster, Shlens, Steiner, Sutskever, Talwar, Tucker, Vanhoucke, Vasudevan, Vi\'{e}gas, Vinyals, Warden, Wattenberg, Wicke, Yu, and Zheng]{tensorflow2015-whitepaper}
Mart\'{i}n Abadi, Ashish Agarwal, Paul Barham, Eugene Brevdo, Zhifeng Chen, Craig Citro, Greg~S. Corrado, Andy Davis, Jeffrey Dean, Matthieu Devin, Sanjay Ghemawat, Ian Goodfellow, Andrew Harp, Geoffrey Irving, Michael Isard, Yangqing Jia, Rafal Jozefowicz, Lukasz Kaiser, Manjunath Kudlur, Josh Levenberg, Dandelion Man\'{e}, Rajat Monga, Sherry Moore, Derek Murray, Chris Olah, Mike Schuster, Jonathon Shlens, Benoit Steiner, Ilya Sutskever, Kunal Talwar, Paul Tucker, Vincent Vanhoucke, Vijay Vasudevan, Fernanda Vi\'{e}gas, Oriol Vinyals, Pete Warden, Martin Wattenberg, Martin Wicke, Yuan Yu, and Xiaoqiang Zheng.
\newblock {TensorFlow}: Large-scale machine learning on heterogeneous systems, 2015.
\newblock URL \url{https://www.tensorflow.org/}.
\newblock Software available from tensorflow.org.

\bibitem[Achiam et~al.(2023)Achiam, Adler, Agarwal, Ahmad, Akkaya, Aleman, Almeida, Altenschmidt, Altman, Anadkat, et~al.]{achiam2023gpt}
Josh Achiam, Steven Adler, Sandhini Agarwal, Lama Ahmad, Ilge Akkaya, Florencia~Leoni Aleman, Diogo Almeida, Janko Altenschmidt, Sam Altman, Shyamal Anadkat, et~al.
\newblock Gpt-4 technical report.
\newblock \emph{arXiv preprint arXiv:2303.08774}, 2023.

\bibitem[Adam et~al.(2019)Adam, Alexandropoulos, Pardalos, and Vrahatis]{adam2019no}
Stavros~P Adam, Stamatios-Aggelos~N Alexandropoulos, Panos~M Pardalos, and Michael~N Vrahatis.
\newblock No free lunch theorem: A review.
\newblock \emph{Approximation and optimization: Algorithms, complexity and applications}, pp.\  57--82, 2019.

\bibitem[Aldaco et~al.(2024)Aldaco, Armstrong, Baruch, Bingham, Chan, Draper, Dwibedi, Finn, Florence, Goodrich, et~al.]{aldaco2024aloha}
Jorge Aldaco, Travis Armstrong, Robert Baruch, Jeff Bingham, Sanky Chan, Kenneth Draper, Debidatta Dwibedi, Chelsea Finn, Pete Florence, Spencer Goodrich, et~al.
\newblock Aloha 2: An enhanced low-cost hardware for bimanual teleoperation.
\newblock \emph{arXiv preprint arXiv:2405.02292}, 2024.

\bibitem[Amadio et~al.(2019)Amadio, Colom{\'e}, and Torras]{amadio2019exploiting}
Fabio Amadio, Adri{\`a} Colom{\'e}, and Carme Torras.
\newblock Exploiting symmetries in reinforcement learning of bimanual robotic tasks.
\newblock \emph{IEEE Robotics and Automation Letters}, 4\penalty0 (2):\penalty0 1838--1845, 2019.

\bibitem[Bao et~al.(2023)Bao, Nie, Xue, Cao, Li, Su, and Zhu]{bao2023all}
Fan Bao, Shen Nie, Kaiwen Xue, Yue Cao, Chongxuan Li, Hang Su, and Jun Zhu.
\newblock All are worth words: A vit backbone for diffusion models.
\newblock In \emph{Proceedings of the IEEE/CVF conference on computer vision and pattern recognition}, pp.\  22669--22679, 2023.

\bibitem[Batinica et~al.(2017)Batinica, Nemec, Ude, Rakovi{\'c}, and Gams]{batinica2017compliant}
Aleksandar Batinica, Bojan Nemec, Ale{\v{s}} Ude, Mirko Rakovi{\'c}, and Andrej Gams.
\newblock Compliant movement primitives in a bimanual setting.
\newblock In \emph{2017 IEEE-RAS 17th International Conference on Humanoid Robotics (Humanoids)}, pp.\  365--371. IEEE, 2017.

\bibitem[Belkhale et~al.(2023)Belkhale, Cui, and Sadigh]{belkhale2023hydra}
Suneel Belkhale, Yuchen Cui, and Dorsa Sadigh.
\newblock Hydra: Hybrid robot actions for imitation learning.
\newblock \emph{arxiv}, 2023.

\bibitem[Brohan et~al.(2022)Brohan, Brown, Carbajal, Chebotar, Dabis, Finn, Gopalakrishnan, Hausman, Herzog, Hsu, et~al.]{brohan2022rt}
Anthony Brohan, Noah Brown, Justice Carbajal, Yevgen Chebotar, Joseph Dabis, Chelsea Finn, Keerthana Gopalakrishnan, Karol Hausman, Alex Herzog, Jasmine Hsu, et~al.
\newblock Rt-1: Robotics transformer for real-world control at scale.
\newblock \emph{arXiv preprint arXiv:2212.06817}, 2022.

\bibitem[Brohan et~al.(2023)Brohan, Brown, Carbajal, Chebotar, Chen, Choromanski, Ding, Driess, Dubey, Finn, et~al.]{brohan2023rt}
Anthony Brohan, Noah Brown, Justice Carbajal, Yevgen Chebotar, Xi~Chen, Krzysztof Choromanski, Tianli Ding, Danny Driess, Avinava Dubey, Chelsea Finn, et~al.
\newblock Rt-2: Vision-language-action models transfer web knowledge to robotic control.
\newblock \emph{arXiv preprint arXiv:2307.15818}, 2023.

\bibitem[Brown et~al.(2020)Brown, Mann, Ryder, Subbiah, Kaplan, Dhariwal, Neelakantan, Shyam, Sastry, Askell, et~al.]{brown2020language}
Tom Brown, Benjamin Mann, Nick Ryder, Melanie Subbiah, Jared~D Kaplan, Prafulla Dhariwal, Arvind Neelakantan, Pranav Shyam, Girish Sastry, Amanda Askell, et~al.
\newblock Language models are few-shot learners.
\newblock \emph{Advances in neural information processing systems}, 33:\penalty0 1877--1901, 2020.

\bibitem[Chen et~al.()Chen, Adebola, and Goldberg]{BerkeleyUR5Website}
Lawrence~Yunliang Chen, Simeon Adebola, and Ken Goldberg.
\newblock Berkeley {UR5} demonstration dataset.
\newblock https://sites.google.com/view/berkeley-ur5/home.

\bibitem[Chen et~al.(2023)Chen, Bahl, and Pathak]{chen2023playfusion}
Lili Chen, Shikhar Bahl, and Deepak Pathak.
\newblock Playfusion: Skill acquisition via diffusion from language-annotated play.
\newblock In \emph{CoRL}, 2023.

\bibitem[Chen et~al.(2019)Chen, Wu, and Han]{chen2019capturing}
Xin Chen, Aming Wu, and Yahong Han.
\newblock Capturing the spatio-temporal continuity for video semantic segmentation.
\newblock \emph{IET Image Processing}, 13\penalty0 (14):\penalty0 2813--2820, 2019.

\bibitem[Chi et~al.(2023)Chi, Feng, Du, Xu, Cousineau, Burchfiel, and Song]{chi2023diffusionpolicy}
Cheng Chi, Siyuan Feng, Yilun Du, Zhenjia Xu, Eric Cousineau, Benjamin Burchfiel, and Shuran Song.
\newblock Diffusion policy: Visuomotor policy learning via action diffusion.
\newblock In \emph{Proceedings of Robotics: Science and Systems (RSS)}, 2023.

\bibitem[Chitnis et~al.(2020)Chitnis, Tulsiani, Gupta, and Gupta]{chitnis2020efficient}
Rohan Chitnis, Shubham Tulsiani, Saurabh Gupta, and Abhinav Gupta.
\newblock Efficient bimanual manipulation using learned task schemas.
\newblock In \emph{2020 IEEE International Conference on Robotics and Automation (ICRA)}, pp.\  1149--1155. IEEE, 2020.

\bibitem[Collaboration et~al.(2023)Collaboration, O'Neill, Rehman, Maddukuri, Gupta, Padalkar, Lee, Pooley, Gupta, Mandlekar, Jain, Tung, Bewley, Herzog, Irpan, Khazatsky, Rai, Gupta, Wang, Kolobov, Singh, Garg, Kembhavi, Xie, Brohan, Raffin, Sharma, Yavary, Jain, Balakrishna, Wahid, Burgess-Limerick, Kim, Schölkopf, Wulfe, Ichter, Lu, Xu, Le, Finn, Wang, Xu, Chi, Huang, Chan, Agia, Pan, Fu, Devin, Xu, Morton, Driess, Chen, Pathak, Shah, Büchler, Jayaraman, Kalashnikov, Sadigh, Johns, Foster, Liu, Ceola, Xia, Zhao, Frujeri, Stulp, Zhou, Sukhatme, Salhotra, Yan, Feng, Schiavi, Berseth, Kahn, Yang, Wang, Su, Fang, Shi, Bao, Amor, Christensen, Furuta, Walke, Fang, Ha, Mordatch, Radosavovic, Leal, Liang, Abou-Chakra, Kim, Drake, Peters, Schneider, Hsu, Bohg, Bingham, Wu, Gao, Hu, Wu, Wu, Sun, Luo, Gu, Tan, Oh, Wu, Lu, Yang, Malik, Silvério, Hejna, Booher, Tompson, Yang, Salvador, Lim, Han, Wang, Rao, Pertsch, Hausman, Go, Gopalakrishnan, Goldberg, Byrne, Oslund, Kawaharazuka, Black, Lin, Zhang, Ehsani,
  Lekkala, Ellis, Rana, Srinivasan, Fang, Singh, Zeng, Hatch, Hsu, Itti, Chen, Pinto, Fei-Fei, Tan, Fan, Ott, Lee, Weihs, Chen, Lepert, Memmel, Tomizuka, Itkina, Castro, Spero, Du, Ahn, Yip, Zhang, Ding, Heo, Srirama, Sharma, Kim, Kanazawa, Hansen, Heess, Joshi, Suenderhauf, Liu, Palo, Shafiullah, Mees, Kroemer, Bastani, Sanketi, Miller, Yin, Wohlhart, Xu, Fagan, Mitrano, Sermanet, Abbeel, Sundaresan, Chen, Vuong, Rafailov, Tian, Doshi, Mart{'i}n-Mart{'i}n, Baijal, Scalise, Hendrix, Lin, Qian, Zhang, Mendonca, Shah, Hoque, Julian, Bustamante, Kirmani, Levine, Lin, Moore, Bahl, Dass, Sonawani, Song, Xu, Haldar, Karamcheti, Adebola, Guist, Nasiriany, Schaal, Welker, Tian, Ramamoorthy, Dasari, Belkhale, Park, Nair, Mirchandani, Osa, Gupta, Harada, Matsushima, Xiao, Kollar, Yu, Ding, Davchev, Zhao, Armstrong, Darrell, Chung, Jain, Vanhoucke, Zhan, Zhou, Burgard, Chen, Chen, Wang, Zhu, Geng, Liu, Liangwei, Li, Pang, Lu, Ma, Kim, Chebotar, Zhou, Zhu, Wu, Xu, Wang, Bisk, Dou, Cho, Lee, Cui, Cao, Wu, Tang, Zhu,
  Zhang, Jiang, Li, Li, Iwasawa, Matsuo, Ma, Xu, Cui, Zhang, Fu, and Lin]{padalkar2023open}
Open X-Embodiment Collaboration, Abby O'Neill, Abdul Rehman, Abhiram Maddukuri, Abhishek Gupta, Abhishek Padalkar, Abraham Lee, Acorn Pooley, Agrim Gupta, Ajay Mandlekar, Ajinkya Jain, Albert Tung, Alex Bewley, Alex Herzog, Alex Irpan, Alexander Khazatsky, Anant Rai, Anchit Gupta, Andrew Wang, Andrey Kolobov, Anikait Singh, Animesh Garg, Aniruddha Kembhavi, Annie Xie, Anthony Brohan, Antonin Raffin, Archit Sharma, Arefeh Yavary, Arhan Jain, Ashwin Balakrishna, Ayzaan Wahid, Ben Burgess-Limerick, Beomjoon Kim, Bernhard Schölkopf, Blake Wulfe, Brian Ichter, Cewu Lu, Charles Xu, Charlotte Le, Chelsea Finn, Chen Wang, Chenfeng Xu, Cheng Chi, Chenguang Huang, Christine Chan, Christopher Agia, Chuer Pan, Chuyuan Fu, Coline Devin, Danfei Xu, Daniel Morton, Danny Driess, Daphne Chen, Deepak Pathak, Dhruv Shah, Dieter Büchler, Dinesh Jayaraman, Dmitry Kalashnikov, Dorsa Sadigh, Edward Johns, Ethan Foster, Fangchen Liu, Federico Ceola, Fei Xia, Feiyu Zhao, Felipe~Vieira Frujeri, Freek Stulp, Gaoyue Zhou, Gaurav~S.
  Sukhatme, Gautam Salhotra, Ge~Yan, Gilbert Feng, Giulio Schiavi, Glen Berseth, Gregory Kahn, Guangwen Yang, Guanzhi Wang, Hao Su, Hao-Shu Fang, Haochen Shi, Henghui Bao, Heni~Ben Amor, Henrik~I Christensen, Hiroki Furuta, Homer Walke, Hongjie Fang, Huy Ha, Igor Mordatch, Ilija Radosavovic, Isabel Leal, Jacky Liang, Jad Abou-Chakra, Jaehyung Kim, Jaimyn Drake, Jan Peters, Jan Schneider, Jasmine Hsu, Jeannette Bohg, Jeffrey Bingham, Jeffrey Wu, Jensen Gao, Jiaheng Hu, Jiajun Wu, Jialin Wu, Jiankai Sun, Jianlan Luo, Jiayuan Gu, Jie Tan, Jihoon Oh, Jimmy Wu, Jingpei Lu, Jingyun Yang, Jitendra Malik, João Silvério, Joey Hejna, Jonathan Booher, Jonathan Tompson, Jonathan Yang, Jordi Salvador, Joseph~J. Lim, Junhyek Han, Kaiyuan Wang, Kanishka Rao, Karl Pertsch, Karol Hausman, Keegan Go, Keerthana Gopalakrishnan, Ken Goldberg, Kendra Byrne, Kenneth Oslund, Kento Kawaharazuka, Kevin Black, Kevin Lin, Kevin Zhang, Kiana Ehsani, Kiran Lekkala, Kirsty Ellis, Krishan Rana, Krishnan Srinivasan, Kuan Fang, Kunal~Pratap
  Singh, Kuo-Hao Zeng, Kyle Hatch, Kyle Hsu, Laurent Itti, Lawrence~Yunliang Chen, Lerrel Pinto, Li~Fei-Fei, Liam Tan, Linxi~"Jim" Fan, Lionel Ott, Lisa Lee, Luca Weihs, Magnum Chen, Marion Lepert, Marius Memmel, Masayoshi Tomizuka, Masha Itkina, Mateo~Guaman Castro, Max Spero, Maximilian Du, Michael Ahn, Michael~C. Yip, Mingtong Zhang, Mingyu Ding, Minho Heo, Mohan~Kumar Srirama, Mohit Sharma, Moo~Jin Kim, Naoaki Kanazawa, Nicklas Hansen, Nicolas Heess, Nikhil~J Joshi, Niko Suenderhauf, Ning Liu, Norman~Di Palo, Nur Muhammad~Mahi Shafiullah, Oier Mees, Oliver Kroemer, Osbert Bastani, Pannag~R Sanketi, Patrick~"Tree" Miller, Patrick Yin, Paul Wohlhart, Peng Xu, Peter~David Fagan, Peter Mitrano, Pierre Sermanet, Pieter Abbeel, Priya Sundaresan, Qiuyu Chen, Quan Vuong, Rafael Rafailov, Ran Tian, Ria Doshi, Roberto Mart{'i}n-Mart{'i}n, Rohan Baijal, Rosario Scalise, Rose Hendrix, Roy Lin, Runjia Qian, Ruohan Zhang, Russell Mendonca, Rutav Shah, Ryan Hoque, Ryan Julian, Samuel Bustamante, Sean Kirmani, Sergey
  Levine, Shan Lin, Sherry Moore, Shikhar Bahl, Shivin Dass, Shubham Sonawani, Shuran Song, Sichun Xu, Siddhant Haldar, Siddharth Karamcheti, Simeon Adebola, Simon Guist, Soroush Nasiriany, Stefan Schaal, Stefan Welker, Stephen Tian, Subramanian Ramamoorthy, Sudeep Dasari, Suneel Belkhale, Sungjae Park, Suraj Nair, Suvir Mirchandani, Takayuki Osa, Tanmay Gupta, Tatsuya Harada, Tatsuya Matsushima, Ted Xiao, Thomas Kollar, Tianhe Yu, Tianli Ding, Todor Davchev, Tony~Z. Zhao, Travis Armstrong, Trevor Darrell, Trinity Chung, Vidhi Jain, Vincent Vanhoucke, Wei Zhan, Wenxuan Zhou, Wolfram Burgard, Xi~Chen, Xiangyu Chen, Xiaolong Wang, Xinghao Zhu, Xinyang Geng, Xiyuan Liu, Xu~Liangwei, Xuanlin Li, Yansong Pang, Yao Lu, Yecheng~Jason Ma, Yejin Kim, Yevgen Chebotar, Yifan Zhou, Yifeng Zhu, Yilin Wu, Ying Xu, Yixuan Wang, Yonatan Bisk, Yongqiang Dou, Yoonyoung Cho, Youngwoon Lee, Yuchen Cui, Yue Cao, Yueh-Hua Wu, Yujin Tang, Yuke Zhu, Yunchu Zhang, Yunfan Jiang, Yunshuang Li, Yunzhu Li, Yusuke Iwasawa, Yutaka Matsuo,
  Zehan Ma, Zhuo Xu, Zichen~Jeff Cui, Zichen Zhang, Zipeng Fu, and Zipeng Lin.
\newblock Open {X-E}mbodiment: Robotic learning datasets and {RT-X} models.
\newblock \url{https://arxiv.org/abs/2310.08864}, 2023.

\bibitem[Colom{\'e} \& Torras(2018)Colom{\'e} and Torras]{colome2018dimensionality}
Adria Colom{\'e} and Carme Torras.
\newblock Dimensionality reduction for dynamic movement primitives and application to bimanual manipulation of clothes.
\newblock \emph{IEEE Transactions on Robotics}, 34\penalty0 (3):\penalty0 602--615, 2018.

\bibitem[Colom{\'e} \& Torras(2020)Colom{\'e} and Torras]{colome2020reinforcement}
Adri{\`a} Colom{\'e} and Carme Torras.
\newblock \emph{Reinforcement learning of bimanual robot skills}.
\newblock Springer, 2020.

\bibitem[Dass et~al.(2023)Dass, Yapeter, Zhang, Zhang, Pertsch, Nikolaidis, and Lim]{dass2023jacoplay}
Shivin Dass, Jullian Yapeter, Jesse Zhang, Jiahui Zhang, Karl Pertsch, Stefanos Nikolaidis, and Joseph~J. Lim.
\newblock Clvr jaco play dataset, 2023.
\newblock URL \url{https://github.com/clvrai/clvr_jaco_play_dataset}.

\bibitem[de~Wit et~al.(2012)de~Wit, Siciliano, and Bastin]{de2012theory}
Carlos~Canudas de~Wit, Bruno Siciliano, and Georges Bastin.
\newblock \emph{Theory of robot control}.
\newblock Springer Science \& Business Media, 2012.

\bibitem[Driess et~al.(2023)Driess, Xia, Sajjadi, Lynch, Chowdhery, Ichter, Wahid, Tompson, Vuong, Yu, et~al.]{driess2023palm}
Danny Driess, Fei Xia, Mehdi~SM Sajjadi, Corey Lynch, Aakanksha Chowdhery, Brian Ichter, Ayzaan Wahid, Jonathan Tompson, Quan Vuong, Tianhe Yu, et~al.
\newblock Palm-e: An embodied multimodal language model.
\newblock \emph{arXiv preprint arXiv:2303.03378}, 2023.

\bibitem[Edsinger \& Kemp(2007)Edsinger and Kemp]{edsinger2007two}
Aaron Edsinger and Charles~C Kemp.
\newblock Two arms are better than one: A behavior based control system for assistive bimanual manipulation.
\newblock In \emph{Recent Progress in Robotics: Viable Robotic Service to Human: An Edition of the Selected Papers from the 13th International Conference on Advanced Robotics}, pp.\  345--355. Springer, 2007.

\bibitem[Fang et~al.(2023)Fang, Fang, Tang, Liu, Wang, Zhu, and Lu]{fang2023rh20t}
Hao-Shu Fang, Hongjie Fang, Zhenyu Tang, Jirong Liu, Junbo Wang, Haoyi Zhu, and Cewu Lu.
\newblock Rh20t: A robotic dataset for learning diverse skills in one-shot.
\newblock In \emph{RSS 2023 Workshop on Learning for Task and Motion Planning}, 2023.

\bibitem[Federico~Ceola(2023)]{ceola2023lhmanip}
Niko~S{\"u}nderhauf Federico~Ceola, Krishan~Rana.
\newblock Lhmanip: A dataset for long horizon manipulation tasks., 2023.

\bibitem[Feng et~al.(2023)Feng, Hansen, Xiong, Rajagopalan, and Wang]{Feng2023Finetuning}
Yunhai Feng, Nicklas Hansen, Ziyan Xiong, Chandramouli Rajagopalan, and Xiaolong Wang.
\newblock Finetuning offline world models in the real world.
\newblock \emph{arXiv preprint arXiv:2310.16029}, 2023.

\bibitem[Figueroa \& Billard(2017)Figueroa and Billard]{figueroa2017learning}
Nadia Figueroa and Aude Billard.
\newblock Learning complex manipulation tasks from heterogeneous and unstructured demonstrations.
\newblock In \emph{Proceedings of Workshop on Synergies between Learning and Interaction}, 2017.

\bibitem[Franzese et~al.(2023)Franzese, de~Souza~Rosa, Verburg, Peternel, and Kober]{franzese2023interactive}
Giovanni Franzese, Leandro de~Souza~Rosa, Tim Verburg, Luka Peternel, and Jens Kober.
\newblock Interactive imitation learning of bimanual movement primitives.
\newblock \emph{IEEE/ASME Transactions on Mechatronics}, 2023.

\bibitem[Fu et~al.(2024)Fu, Zhao, and Finn]{fu2024mobile}
Zipeng Fu, Tony~Z Zhao, and Chelsea Finn.
\newblock Mobile aloha: Learning bimanual mobile manipulation with low-cost whole-body teleoperation.
\newblock \emph{arXiv preprint arXiv:2401.02117}, 2024.

\bibitem[Ghosh et~al.(2023)Ghosh, Walke, Pertsch, Black, Mees, Dasari, Hejna, Xu, Luo, et~al.]{team2023octo}
Dibya Ghosh, Homer Walke, Karl Pertsch, Kevin Black, Oier Mees, Sudeep Dasari, Joey Hejna, Charles Xu, Jianlan Luo, et~al.
\newblock Octo: An open-source generalist robot policy, 2023.

\bibitem[Grannen et~al.(2023{\natexlab{a}})Grannen, Wu, Belkhale, and Sadigh]{grannen2023learning}
Jennifer Grannen, Yilin Wu, Suneel Belkhale, and Dorsa Sadigh.
\newblock Learning bimanual scooping policies for food acquisition.
\newblock In \emph{Conference on Robot Learning}, pp.\  1510--1519. PMLR, 2023{\natexlab{a}}.

\bibitem[Grannen et~al.(2023{\natexlab{b}})Grannen, Wu, Vu, and Sadigh]{grannen2023stabilize}
Jennifer Grannen, Yilin Wu, Brandon Vu, and Dorsa Sadigh.
\newblock Stabilize to act: Learning to coordinate for bimanual manipulation.
\newblock In \emph{Conference on Robot Learning}, pp.\  563--576. PMLR, 2023{\natexlab{b}}.

\bibitem[Grotz et~al.(2024)Grotz, Shridhar, Asfour, and Fox]{grotz2024peract2}
Markus Grotz, Mohit Shridhar, Tamim Asfour, and Dieter Fox.
\newblock Peract2: A perceiver actor framework for bimanual manipulation tasks.
\newblock \emph{arXiv preprint arXiv:2407.00278}, 2024.

\bibitem[Gu et~al.(2023)Gu, Xiang, Li, Ling, Liu, Mu, Tang, Tao, Wei, Yao, Yuan, Xie, Huang, Chen, and Su]{gu2023maniskill2}
Jiayuan Gu, Fanbo Xiang, Xuanlin Li, Zhan Ling, Xiqiang Liu, Tongzhou Mu, Yihe Tang, Stone Tao, Xinyue Wei, Yunchao Yao, Xiaodi Yuan, Pengwei Xie, Zhiao Huang, Rui Chen, and Hao Su.
\newblock Maniskill2: A unified benchmark for generalizable manipulation skills.
\newblock In \emph{International Conference on Learning Representations}, 2023.

\bibitem[Hendrycks \& Gimpel(2016)Hendrycks and Gimpel]{hendrycks2016gaussian}
Dan Hendrycks and Kevin Gimpel.
\newblock Gaussian error linear units (gelus).
\newblock \emph{arXiv preprint arXiv:1606.08415}, 2016.

\bibitem[Henry et~al.(2020)Henry, Dachapally, Pawar, and Chen]{henry2020query}
Alex Henry, Prudhvi~Raj Dachapally, Shubham Pawar, and Yuxuan Chen.
\newblock Query-key normalization for transformers.
\newblock \emph{arXiv preprint arXiv:2010.04245}, 2020.

\bibitem[Heo et~al.(2023)Heo, Lee, Lee, and Lim]{heo2023furniturebench}
Minho Heo, Youngwoon Lee, Doohyun Lee, and Joseph~J. Lim.
\newblock Furniturebench: Reproducible real-world benchmark for long-horizon complex manipulation.
\newblock In \emph{Robotics: Science and Systems}, 2023.

\bibitem[Ho et~al.(2020)Ho, Jain, and Abbeel]{ho2020denoising}
Jonathan Ho, Ajay Jain, and Pieter Abbeel.
\newblock Denoising diffusion probabilistic models.
\newblock \emph{Advances in neural information processing systems}, 33:\penalty0 6840--6851, 2020.

\bibitem[Hu et~al.(2021)Hu, Shen, Wallis, Allen-Zhu, Li, Wang, Wang, and Chen]{hu2021lora}
Edward~J Hu, Yelong Shen, Phillip Wallis, Zeyuan Allen-Zhu, Yuanzhi Li, Shean Wang, Lu~Wang, and Weizhu Chen.
\newblock Lora: Low-rank adaptation of large language models.
\newblock \emph{arXiv preprint arXiv:2106.09685}, 2021.

\bibitem[Huang et~al.(2024)Huang, K{\"u}mmerle, and Zhang]{huang2024unitnorm}
Nan Huang, Christian K{\"u}mmerle, and Xiang Zhang.
\newblock Unitnorm: Rethinking normalization for transformers in time series.
\newblock \emph{arXiv preprint arXiv:2405.15903}, 2024.

\bibitem[Jang et~al.(2021)Jang, Irpan, Khansari, Kappler, Ebert, Lynch, Levine, and Finn]{jang2021bc}
Eric Jang, Alex Irpan, Mohi Khansari, Daniel Kappler, Frederik Ebert, Corey Lynch, Sergey Levine, and Chelsea Finn.
\newblock {BC}-z: Zero-shot task generalization with robotic imitation learning.
\newblock In \emph{5th Annual Conference on Robot Learning}, 2021.
\newblock URL \url{https://openreview.net/forum?id=8kbp23tSGYv}.

\bibitem[Jia et~al.(2024)Jia, Blessing, Jiang, Reuss, Donat, Lioutikov, and Neumann]{jia2024towards}
Xiaogang Jia, Denis Blessing, Xinkai Jiang, Moritz Reuss, Atalay Donat, Rudolf Lioutikov, and Gerhard Neumann.
\newblock Towards diverse behaviors: A benchmark for imitation learning with human demonstrations.
\newblock \emph{arXiv preprint arXiv:2402.14606}, 2024.

\bibitem[Kalashnikov et~al.(2018)Kalashnikov, Irpan, Pastor, Ibarz, Herzog, Jang, Quillen, Holly, Kalakrishnan, Vanhoucke, et~al.]{kalashnikov2018qt}
Dmitry Kalashnikov, Alex Irpan, Peter Pastor, Julian Ibarz, Alexander Herzog, Eric Jang, Deirdre Quillen, Ethan Holly, Mrinal Kalakrishnan, Vincent Vanhoucke, et~al.
\newblock Qt-opt: Scalable deep reinforcement learning for vision-based robotic manipulation.
\newblock \emph{arXiv preprint arXiv:1806.10293}, 2018.

\bibitem[Khazatsky et~al.(2024)Khazatsky, Pertsch, Nair, Balakrishna, Dasari, Karamcheti, Nasiriany, Srirama, Chen, Ellis, et~al.]{khazatsky2024droid}
Alexander Khazatsky, Karl Pertsch, Suraj Nair, Ashwin Balakrishna, Sudeep Dasari, Siddharth Karamcheti, Soroush Nasiriany, Mohan~Kumar Srirama, Lawrence~Yunliang Chen, Kirsty Ellis, et~al.
\newblock Droid: A large-scale in-the-wild robot manipulation dataset.
\newblock \emph{arXiv preprint arXiv:2403.12945}, 2024.

\bibitem[Kim et~al.(2023)Kim, Han, Kim, and Kim]{kimpre}
Minchan Kim, Junhyek Han, Jaehyung Kim, and Beomjoon Kim.
\newblock Pre-and post-contact policy decomposition for non-prehensile manipulation with zero-shot sim-to-real transfer.
\newblock 2023.

\bibitem[Kim et~al.(2024)Kim, Pertsch, Karamcheti, Xiao, Balakrishna, Nair, Rafailov, Foster, Lam, Sanketi, et~al.]{kim2024openvla}
Moo~Jin Kim, Karl Pertsch, Siddharth Karamcheti, Ted Xiao, Ashwin Balakrishna, Suraj Nair, Rafael Rafailov, Ethan Foster, Grace Lam, Pannag Sanketi, et~al.
\newblock Openvla: An open-source vision-language-action model.
\newblock \emph{arXiv preprint arXiv:2406.09246}, 2024.

\bibitem[Kirillov et~al.(2023)Kirillov, Mintun, Ravi, Mao, Rolland, Gustafson, Xiao, Whitehead, Berg, Lo, et~al.]{kirillov2023segment}
Alexander Kirillov, Eric Mintun, Nikhila Ravi, Hanzi Mao, Chloe Rolland, Laura Gustafson, Tete Xiao, Spencer Whitehead, Alexander~C Berg, Wan-Yen Lo, et~al.
\newblock Segment anything.
\newblock In \emph{Proceedings of the IEEE/CVF International Conference on Computer Vision}, pp.\  4015--4026, 2023.

\bibitem[Kouvaritakis \& Cannon(2016)Kouvaritakis and Cannon]{kouvaritakis2016model}
Basil Kouvaritakis and Mark Cannon.
\newblock Model predictive control.
\newblock \emph{Switzerland: Springer International Publishing}, 38:\penalty0 13--56, 2016.

\bibitem[Krebs et~al.(2021)Krebs, Meixner, Patzer, and Asfour]{krebs2021kit}
Franziska Krebs, Andre Meixner, Isabel Patzer, and Tamim Asfour.
\newblock The kit bimanual manipulation dataset.
\newblock In \emph{2020 IEEE-RAS 20th International Conference on Humanoid Robots (Humanoids)}, pp.\  499--506. IEEE, 2021.

\bibitem[Kumar et~al.(2024)Kumar, Shah, Zhou, Moens, Caggiano, Gupta, and Rajeswaran]{RoboHive}
Vikash Kumar, Rutav Shah, Gaoyue Zhou, Vincent Moens, Vittorio Caggiano, Abhishek Gupta, and Aravind Rajeswaran.
\newblock Robohive: A unified framework for robot learning.
\newblock \emph{Advances in Neural Information Processing Systems}, 36, 2024.

\bibitem[Lee et~al.(2019)Lee, Zhu, Srinivasan, Shah, Savarese, Fei-Fei, Garg, and Bohg]{lee2019icra}
Michelle~A Lee, Yuke Zhu, Krishnan Srinivasan, Parth Shah, Silvio Savarese, Li~Fei-Fei, Animesh Garg, and Jeannette Bohg.
\newblock Making sense of vision and touch: Self-supervised learning of multimodal representations for contact-rich tasks.
\newblock In \emph{2019 IEEE International Conference on Robotics and Automation (ICRA)}, 2019.
\newblock URL \url{https://arxiv.org/abs/1810.10191}.

\bibitem[Li(2006)]{li2006optimal}
Weiwei Li.
\newblock \emph{Optimal control for biological movement systems}.
\newblock University of California, San Diego, 2006.

\bibitem[Liang et~al.(2022)Liang, Quader, Chi, Chen, Dai, Lu, and Wang]{liang2022self}
Hanwen Liang, Niamul Quader, Zhixiang Chi, Lizhe Chen, Peng Dai, Juwei Lu, and Yang Wang.
\newblock Self-supervised spatiotemporal representation learning by exploiting video continuity.
\newblock In \emph{Proceedings of the AAAI Conference on Artificial Intelligence}, volume~36, pp.\  1564--1573, 2022.

\bibitem[Lioutikov et~al.(2016)Lioutikov, Kroemer, Maeda, and Peters]{lioutikov2016learning}
Rudolf Lioutikov, Oliver Kroemer, Guilherme Maeda, and Jan Peters.
\newblock Learning manipulation by sequencing motor primitives with a two-armed robot.
\newblock In \emph{Intelligent Autonomous Systems 13: Proceedings of the 13th International Conference IAS-13}, pp.\  1601--1611. Springer, 2016.

\bibitem[Liu et~al.(2023)Liu, Nasiriany, Zhang, Bao, and Zhu]{liu2022robot}
Huihan Liu, Soroush Nasiriany, Lance Zhang, Zhiyao Bao, and Yuke Zhu.
\newblock Robot learning on the job: Human-in-the-loop autonomy and learning during deployment.
\newblock In \emph{Robotics: Science and Systems (RSS)}, 2023.

\bibitem[Liu et~al.(2024)Liu, Arthur, He, Seita, and Sukhatme]{liu2024voxact}
I~Liu, Chun Arthur, Sicheng He, Daniel Seita, and Gaurav Sukhatme.
\newblock Voxact-b: Voxel-based acting and stabilizing policy for bimanual manipulation.
\newblock \emph{arXiv preprint arXiv:2407.04152}, 2024.

\bibitem[Liu et~al.(2022)Liu, Wang, Zhang, and Sun]{liu2022petr}
Yingfei Liu, Tiancai Wang, Xiangyu Zhang, and Jian Sun.
\newblock Petr: Position embedding transformation for multi-view 3d object detection.
\newblock In \emph{European Conference on Computer Vision}, pp.\  531--548. Springer, 2022.

\bibitem[Lu et~al.(2022)Lu, Zhou, Bao, Chen, Li, and Zhu]{lu2022dpm}
Cheng Lu, Yuhao Zhou, Fan Bao, Jianfei Chen, Chongxuan Li, and Jun Zhu.
\newblock Dpm-solver++: Fast solver for guided sampling of diffusion probabilistic models.
\newblock \emph{arXiv preprint arXiv:2211.01095}, 2022.

\bibitem[Luo et~al.(2023)Luo, Xu, Geng, Feng, Fang, Tan, Schaal, and Levine]{luo2023multistage}
Jianlan Luo, Charles Xu, Xinyang Geng, Gilbert Feng, Kuan Fang, Liam Tan, Stefan Schaal, and Sergey Levine.
\newblock Multi-stage cable routing through hierarchical imitation learning.
\newblock \emph{arXiv pre-print}, 2023.
\newblock URL \url{https://arxiv.org/abs/2307.08927}.

\bibitem[Luo et~al.(2024)Luo, Xu, Liu, Tan, Lin, Wu, Abbeel, and Levine]{luo2024fmbfunctionalmanipulationbenchmark}
Jianlan Luo, Charles Xu, Fangchen Liu, Liam Tan, Zipeng Lin, Jeffrey Wu, Pieter Abbeel, and Sergey Levine.
\newblock Fmb: a functional manipulation benchmark for generalizable robotic learning, 2024.
\newblock URL \url{https://arxiv.org/abs/2401.08553}.

\bibitem[Lynch et~al.(2022)Lynch, Wahid, Tompson, Ding, Betker, Baruch, Armstrong, and Florence]{lynch2022interactivelanguagetalkingrobots}
Corey Lynch, Ayzaan Wahid, Jonathan Tompson, Tianli Ding, James Betker, Robert Baruch, Travis Armstrong, and Pete Florence.
\newblock Interactive language: Talking to robots in real time, 2022.
\newblock URL \url{https://arxiv.org/abs/2210.06407}.

\bibitem[Matsushima et~al.(2023)Matsushima, Furuta, Iwasawa, and Matsuo]{matsushima2023weblab}
Tatsuya Matsushima, Hiroki Furuta, Yusuke Iwasawa, and Yutaka Matsuo.
\newblock Weblab xarm dataset, 2023.

\bibitem[Mees et~al.(2022)Mees, Hermann, Rosete-Beas, and Burgard]{mees2022calvin}
Oier Mees, Lukas Hermann, Erick Rosete-Beas, and Wolfram Burgard.
\newblock Calvin: A benchmark for language-conditioned policy learning for long-horizon robot manipulation tasks.
\newblock \emph{IEEE Robotics and Automation Letters (RA-L)}, 7\penalty0 (3):\penalty0 7327--7334, 2022.

\bibitem[Mirrazavi~Salehian et~al.(2017)Mirrazavi~Salehian, Figueroa~Fernandez, and Billard]{mirrazavi2017dynamical}
Seyed~Sina Mirrazavi~Salehian, Nadia~Barbara Figueroa~Fernandez, and Aude Billard.
\newblock Dynamical system-based motion planning for multi-arm systems: Reaching for moving objects.
\newblock In \emph{IJCAI'17: Proceedings of the 26th International Joint Conference on Artificial Intelligence}, pp.\  4914--4918, 2017.

\bibitem[Nasiriany et~al.(2022)Nasiriany, Gao, Mandlekar, and Zhu]{nasiriany2022sailor}
Soroush Nasiriany, Tian Gao, Ajay Mandlekar, and Yuke Zhu.
\newblock Learning and retrieval from prior data for skill-based imitation learning.
\newblock In \emph{Conference on Robot Learning (CoRL)}, 2022.

\bibitem[Nichol \& Dhariwal(2021)Nichol and Dhariwal]{nichol2021improved}
Alexander~Quinn Nichol and Prafulla Dhariwal.
\newblock Improved denoising diffusion probabilistic models.
\newblock In \emph{International conference on machine learning}, pp.\  8162--8171. PMLR, 2021.

\bibitem[Oh et~al.(2023)Oh, Kanazawa, and Kawaharazuka]{oh2023pr2utokyodatasets}
Jihoon Oh, Naoaki Kanazawa, and Kento Kawaharazuka.
\newblock X-embodiment u-tokyo pr2 datasets, 2023.
\newblock URL \url{https://github.com/ojh6404/rlds_dataset_builder}.

\bibitem[Osa(2022)]{Osa22}
Takayuki Osa.
\newblock Motion planning by learning the solution manifold in trajectory optimization.
\newblock \emph{The International Journal of Robotics Research}, 41\penalty0 (3):\penalty0 291--311, 2022.

\bibitem[Padalkar et~al.(2023)Padalkar, Quere, Raffin, Silv{\'e}rio, and Stulp]{padalkar2023guided}
Abhishek Padalkar, Gabriel Quere, Antonin Raffin, Jo{\~a}o Silv{\'e}rio, and Freek Stulp.
\newblock A guided reinforcement learning approach using shared control templates for learning manipulation skills in the real world.
\newblock \emph{Research square preprint rs-3289569/v1}, 2023.

\bibitem[Pan \& Yang(2009)Pan and Yang]{pan2009survey}
Sinno~Jialin Pan and Qiang Yang.
\newblock A survey on transfer learning.
\newblock \emph{IEEE Transactions on knowledge and data engineering}, 22\penalty0 (10):\penalty0 1345--1359, 2009.

\bibitem[Pari et~al.(2021)Pari, Shafiullah, Arunachalam, and Pinto]{pari2021surprising}
Jyothish Pari, Nur~Muhammad Shafiullah, Sridhar~Pandian Arunachalam, and Lerrel Pinto.
\newblock The surprising effectiveness of representation learning for visual imitation, 2021.

\bibitem[Paszke et~al.(2019)Paszke, Gross, Massa, Lerer, Bradbury, Chanan, Killeen, Lin, Gimelshein, Antiga, et~al.]{paszke2019pytorch}
Adam Paszke, Sam Gross, Francisco Massa, Adam Lerer, James Bradbury, Gregory Chanan, Trevor Killeen, Zeming Lin, Natalia Gimelshein, Luca Antiga, et~al.
\newblock Pytorch: An imperative style, high-performance deep learning library.
\newblock \emph{Advances in neural information processing systems}, 32, 2019.

\bibitem[Pearce et~al.(2023)Pearce, Rashid, Kanervisto, Bignell, Sun, Georgescu, Macua, Tan, Momennejad, Hofmann, and Devlin]{pearce2023imitating}
Tim Pearce, Tabish Rashid, Anssi Kanervisto, Dave Bignell, Mingfei Sun, Raluca Georgescu, Sergio~Valcarcel Macua, Shan~Zheng Tan, Ida Momennejad, Katja Hofmann, and Sam Devlin.
\newblock Imitating human behaviour with diffusion models.
\newblock In \emph{The Eleventh International Conference on Learning Representations}, 2023.

\bibitem[Peebles \& Xie(2023)Peebles and Xie]{peebles2023scalable}
William Peebles and Saining Xie.
\newblock Scalable diffusion models with transformers.
\newblock In \emph{Proceedings of the IEEE/CVF International Conference on Computer Vision}, pp.\  4195--4205, 2023.

\bibitem[Radford et~al.(2018)Radford, Narasimhan, Salimans, Sutskever, et~al.]{radford2018improving}
Alec Radford, Karthik Narasimhan, Tim Salimans, Ilya Sutskever, et~al.
\newblock Improving language understanding by generative pre-training.
\newblock 2018.

\bibitem[Radford et~al.(2021)Radford, Kim, Hallacy, Ramesh, Goh, Agarwal, Sastry, Askell, Mishkin, Clark, et~al.]{radford2021learning}
Alec Radford, Jong~Wook Kim, Chris Hallacy, Aditya Ramesh, Gabriel Goh, Sandhini Agarwal, Girish Sastry, Amanda Askell, Pamela Mishkin, Jack Clark, et~al.
\newblock Learning transferable visual models from natural language supervision.
\newblock In \emph{International conference on machine learning}, pp.\  8748--8763. PMLR, 2021.

\bibitem[Radosavovic et~al.(2022)Radosavovic, Xiao, James, Abbeel, Malik, and Darrell]{Radosavovic2022}
Ilija Radosavovic, Tete Xiao, Stephen James, Pieter Abbeel, Jitendra Malik, and Trevor Darrell.
\newblock Real-world robot learning with masked visual pre-training.
\newblock In \emph{CoRL}, 2022.

\bibitem[Raffel et~al.(2020)Raffel, Shazeer, Roberts, Lee, Narang, Matena, Zhou, Li, and Liu]{2020t5}
Colin Raffel, Noam Shazeer, Adam Roberts, Katherine Lee, Sharan Narang, Michael Matena, Yanqi Zhou, Wei Li, and Peter~J. Liu.
\newblock Exploring the limits of transfer learning with a unified text-to-text transformer.
\newblock \emph{Journal of Machine Learning Research}, 21\penalty0 (140):\penalty0 1--67, 2020.
\newblock URL \url{http://jmlr.org/papers/v21/20-074.html}.

\bibitem[Rakita et~al.(2019)Rakita, Mutlu, Gleicher, and Hiatt]{rakita2019shared}
Daniel Rakita, Bilge Mutlu, Michael Gleicher, and Laura~M Hiatt.
\newblock Shared control--based bimanual robot manipulation.
\newblock \emph{Science Robotics}, 4\penalty0 (30):\penalty0 eaaw0955, 2019.

\bibitem[Rasley et~al.(2020)Rasley, Rajbhandari, Ruwase, and He]{rasley2020deepspeed}
Jeff Rasley, Samyam Rajbhandari, Olatunji Ruwase, and Yuxiong He.
\newblock Deepspeed: System optimizations enable training deep learning models with over 100 billion parameters.
\newblock In \emph{Proceedings of the 26th ACM SIGKDD International Conference on Knowledge Discovery \& Data Mining}, pp.\  3505--3506, 2020.

\bibitem[RethinkRobotics()]{imperialcollege_sawyer_wrist_cam}
RethinkRobotics.
\newblock Sawyer performing table top manipulation.
\newblock \url{https://github.com/RethinkRobotics/sawyer_robot}.

\bibitem[Rosete-Beas et~al.(2022)Rosete-Beas, Mees, Kalweit, Boedecker, and Burgard]{rosete2022tacorl}
Erick Rosete-Beas, Oier Mees, Gabriel Kalweit, Joschka Boedecker, and Wolfram Burgard.
\newblock Latent plans for task agnostic offline reinforcement learning.
\newblock 2022.

\bibitem[Ross et~al.(2011)Ross, Gordon, and Bagnell]{ross2011reduction}
St{\'e}phane Ross, Geoffrey Gordon, and Drew Bagnell.
\newblock A reduction of imitation learning and structured prediction to no-regret online learning.
\newblock In \emph{Proceedings of the fourteenth international conference on artificial intelligence and statistics}, pp.\  627--635. JMLR Workshop and Conference Proceedings, 2011.

\bibitem[Saxena et~al.(2023)Saxena, Sharma, and Kroemer]{saxena2023multiresolution}
Saumya Saxena, Mohit Sharma, and Oliver Kroemer.
\newblock Multi-resolution sensing for real-time control with vision-language models.
\newblock In \emph{7th Annual Conference on Robot Learning}, 2023.
\newblock URL \url{https://openreview.net/forum?id=WuBv9-IGDUA}.

\bibitem[Shafiullah et~al.(2023)Shafiullah, Rai, Etukuru, Liu, Misra, Chintala, and Pinto]{shafiullah2023bringing}
Nur Muhammad~Mahi Shafiullah, Anant Rai, Haritheja Etukuru, Yiqian Liu, Ishan Misra, Soumith Chintala, and Lerrel Pinto.
\newblock On bringing robots home.
\newblock \emph{arXiv preprint arXiv:2311.16098}, 2023.

\bibitem[Shah et~al.(2023{\natexlab{a}})Shah, Sridhar, Bhorkar, Hirose, and Levine]{shah2023gnm}
Dhruv Shah, Ajay Sridhar, Arjun Bhorkar, Noriaki Hirose, and Sergey Levine.
\newblock Gnm: A general navigation model to drive any robot.
\newblock In \emph{2023 IEEE International Conference on Robotics and Automation (ICRA)}, pp.\  7226--7233. IEEE, 2023{\natexlab{a}}.

\bibitem[Shah et~al.(2023{\natexlab{b}})Shah, Mart{\'\i}n-Mart{\'\i}n, and Zhu]{shah2023mutex}
Rutav Shah, Roberto Mart{\'\i}n-Mart{\'\i}n, and Yuke Zhu.
\newblock {MUTEX}: Learning unified policies from multimodal task specifications.
\newblock In \emph{7th Annual Conference on Robot Learning}, 2023{\natexlab{b}}.
\newblock URL \url{https://openreview.net/forum?id=PwqiqaaEzJ}.

\bibitem[Sharma et~al.(2018)Sharma, Mohan, Pinto, and Gupta]{sharma2018multiple}
Pratyusha Sharma, Lekha Mohan, Lerrel Pinto, and Abhinav Gupta.
\newblock Multiple interactions made easy (mime): Large scale demonstrations data for imitation.
\newblock In \emph{Conference on robot learning}, pp.\  906--915. PMLR, 2018.

\bibitem[Shi et~al.(2023)Shi, Xu, Clarke, Li, and Wu]{shi2023robocook}
Haochen Shi, Huazhe Xu, Samuel Clarke, Yunzhu Li, and Jiajun Wu.
\newblock Robocook: Long-horizon elasto-plastic object manipulation with diverse tools.
\newblock \emph{arXiv preprint arXiv:2306.14447}, 2023.

\bibitem[Smith et~al.(2012)Smith, Karayiannidis, Nalpantidis, Gratal, Qi, Dimarogonas, and Kragic]{smith2012dual}
Christian Smith, Yiannis Karayiannidis, Lazaros Nalpantidis, Xavi Gratal, Peng Qi, Dimos~V Dimarogonas, and Danica Kragic.
\newblock Dual arm manipulation—a survey.
\newblock \emph{Robotics and Autonomous systems}, 60\penalty0 (10):\penalty0 1340--1353, 2012.

\bibitem[Sohn et~al.(2015)Sohn, Lee, and Yan]{sohn2015learning}
Kihyuk Sohn, Honglak Lee, and Xinchen Yan.
\newblock Learning structured output representation using deep conditional generative models.
\newblock \emph{Advances in neural information processing systems}, 28, 2015.

\bibitem[Stepputtis et~al.(2020)Stepputtis, Campbell, Phielipp, Lee, Baral, and Ben~Amor]{stepputtis2020language}
Simon Stepputtis, Joseph Campbell, Mariano Phielipp, Stefan Lee, Chitta Baral, and Heni Ben~Amor.
\newblock Language-conditioned imitation learning for robot manipulation tasks.
\newblock \emph{Advances in Neural Information Processing Systems}, 33:\penalty0 13139--13150, 2020.

\bibitem[Stepputtis et~al.(2022)Stepputtis, Bandari, Schaal, and Amor]{stepputtis2022system}
Simon Stepputtis, Maryam Bandari, Stefan Schaal, and Heni~Ben Amor.
\newblock A system for imitation learning of contact-rich bimanual manipulation policies.
\newblock In \emph{2022 IEEE/RSJ International Conference on Intelligent Robots and Systems (IROS)}, pp.\  11810--11817. IEEE, 2022.

\bibitem[Tancik et~al.(2020)Tancik, Srinivasan, Mildenhall, Fridovich-Keil, Raghavan, Singhal, Ramamoorthi, Barron, and Ng]{tancik2020fourier}
Matthew Tancik, Pratul Srinivasan, Ben Mildenhall, Sara Fridovich-Keil, Nithin Raghavan, Utkarsh Singhal, Ravi Ramamoorthi, Jonathan Barron, and Ren Ng.
\newblock Fourier features let networks learn high frequency functions in low dimensional domains.
\newblock \emph{Advances in neural information processing systems}, 33:\penalty0 7537--7547, 2020.

\bibitem[Touvron et~al.(2023)Touvron, Lavril, Izacard, Martinet, Lachaux, Lacroix, Rozi{\`e}re, Goyal, Hambro, Azhar, et~al.]{touvron2023llama}
Hugo Touvron, Thibaut Lavril, Gautier Izacard, Xavier Martinet, Marie-Anne Lachaux, Timoth{\'e}e Lacroix, Baptiste Rozi{\`e}re, Naman Goyal, Eric Hambro, Faisal Azhar, et~al.
\newblock Llama: Open and efficient foundation language models.
\newblock \emph{arXiv preprint arXiv:2302.13971}, 2023.

\bibitem[Walke et~al.(2023)Walke, Black, Lee, Kim, Du, Zheng, Zhao, Hansen-Estruch, Vuong, He, Myers, Fang, Finn, and Levine]{walke2023bridgedata}
Homer Walke, Kevin Black, Abraham Lee, Moo~Jin Kim, Max Du, Chongyi Zheng, Tony Zhao, Philippe Hansen-Estruch, Quan Vuong, Andre He, Vivek Myers, Kuan Fang, Chelsea Finn, and Sergey Levine.
\newblock Bridgedata v2: A dataset for robot learning at scale.
\newblock In \emph{Conference on Robot Learning (CoRL)}, 2023.

\bibitem[Wang et~al.(2024)Wang, Zheng, Nie, Xu, Wang, Ye, Li, Zhang, Cheng, Dong, et~al.]{wang2024all}
Zhiqiang Wang, Hao Zheng, Yunshuang Nie, Wenjun Xu, Qingwei Wang, Hua Ye, Zhe Li, Kaidong Zhang, Xuewen Cheng, Wanxi Dong, et~al.
\newblock All robots in one: A new standard and unified dataset for versatile, general-purpose embodied agents.
\newblock \emph{arXiv preprint arXiv:2408.10899}, 2024.

\bibitem[Wu et~al.(2024)Wu, Yang, Dong, Xie, Su, and Zhu]{wu2024embodied}
Lingxuan Wu, Xiao Yang, Yinpeng Dong, Liuwei Xie, Hang Su, and Jun Zhu.
\newblock Embodied active defense: Leveraging recurrent feedback to counter adversarial patches.
\newblock \emph{arXiv preprint arXiv:2404.00540}, 2024.

\bibitem[Xie et~al.(2020)Xie, Chowdhury, De~Paolis~Kaluza, Zhao, Wong, and Yu]{xie2020deep}
Fan Xie, Alexander Chowdhury, M~De~Paolis~Kaluza, Linfeng Zhao, Lawson Wong, and Rose Yu.
\newblock Deep imitation learning for bimanual robotic manipulation.
\newblock \emph{Advances in neural information processing systems}, 33:\penalty0 2327--2337, 2020.

\bibitem[Yan et~al.(2023)Yan, Wu, and Wang]{ucsd_kitchens}
Ge~Yan, Kris Wu, and Xiaolong Wang.
\newblock {ucsd kitchens Dataset}.
\newblock August 2023.

\bibitem[Yang et~al.(2023)Yang, Sadigh, and Finn]{yang2023polybot}
Jonathan Yang, Dorsa Sadigh, and Chelsea Finn.
\newblock Polybot: Training one policy across robots while embracing variability.
\newblock \emph{arXiv preprint arXiv:2307.03719}, 2023.

\bibitem[Yang et~al.(2024)Yang, Glossop, Bhorkar, Shah, Vuong, Finn, Sadigh, and Levine]{yang2024pushing}
Jonathan Yang, Catherine Glossop, Arjun Bhorkar, Dhruv Shah, Quan Vuong, Chelsea Finn, Dorsa Sadigh, and Sergey Levine.
\newblock Pushing the limits of cross-embodiment learning for manipulation and navigation.
\newblock \emph{arXiv preprint arXiv:2402.19432}, 2024.

\bibitem[Zhai et~al.(2023)Zhai, Mustafa, Kolesnikov, and Beyer]{zhai2023sigmoid}
Xiaohua Zhai, Basil Mustafa, Alexander Kolesnikov, and Lucas Beyer.
\newblock Sigmoid loss for language image pre-training, 2023.

\bibitem[Zhang \& Sennrich(2019)Zhang and Sennrich]{zhang2019root}
Biao Zhang and Rico Sennrich.
\newblock Root mean square layer normalization.
\newblock \emph{Advances in Neural Information Processing Systems}, 32, 2019.

\bibitem[Zhao et~al.(2023)Zhao, Kumar, Levine, and Finn]{zhao2023learning}
Tony~Z Zhao, Vikash Kumar, Sergey Levine, and Chelsea Finn.
\newblock Learning fine-grained bimanual manipulation with low-cost hardware.
\newblock \emph{arXiv preprint arXiv:2304.13705}, 2023.

\bibitem[Zhou et~al.(2023)Zhou, Dean, Srirama, Rajeswaran, Pari, Hatch, Jain, Yu, Abbeel, Pinto, Finn, and Gupta]{zhou2023train}
Gaoyue Zhou, Victoria Dean, Mohan~Kumar Srirama, Aravind Rajeswaran, Jyothish Pari, Kyle Hatch, Aryan Jain, Tianhe Yu, Pieter Abbeel, Lerrel Pinto, Chelsea Finn, and Abhinav Gupta.
\newblock Train offline, test online: A real robot learning benchmark.
\newblock In \emph{2023 IEEE International Conference on Robotics and Automation (ICRA)}, 2023.

\bibitem[Zhou et~al.(2019)Zhou, Barnes, Lu, Yang, and Li]{8953486}
Yi~Zhou, Connelly Barnes, Jingwan Lu, Jimei Yang, and Hao Li.
\newblock On the continuity of rotation representations in neural networks.
\newblock In \emph{2019 IEEE/CVF Conference on Computer Vision and Pattern Recognition (CVPR)}, pp.\  5738--5746, 2019.
\newblock \doi{10.1109/CVPR.2019.00589}.

\bibitem[Zhu et~al.(2023)Zhu, Tian, Xu, Ding, Zhan, and Tomizuka]{fanuc_manipulation2023}
Xinghao Zhu, Ran Tian, Chenfeng Xu, Mingyu Ding, Wei Zhan, and Masayoshi Tomizuka.
\newblock Fanuc manipulation: A dataset for learning-based manipulation with fanuc mate 200id robot.
\newblock 2023.

\bibitem[Zhu et~al.(2022{\natexlab{a}})Zhu, Joshi, Stone, and Zhu]{zhu2022viola}
Yifeng Zhu, Abhishek Joshi, Peter Stone, and Yuke Zhu.
\newblock Viola: Imitation learning for vision-based manipulation with object proposal priors.
\newblock \emph{6th Annual Conference on Robot Learning (CoRL)}, 2022{\natexlab{a}}.

\bibitem[Zhu et~al.(2022{\natexlab{b}})Zhu, Stone, and Zhu]{zhu2022bottom}
Yifeng Zhu, Peter Stone, and Yuke Zhu.
\newblock Bottom-up skill discovery from unsegmented demonstrations for long-horizon robot manipulation.
\newblock \emph{IEEE Robotics and Automation Letters}, 7\penalty0 (2):\penalty0 4126--4133, 2022{\natexlab{b}}.

\bibitem[Ziegler et~al.(2019)Ziegler, Stiennon, Wu, Brown, Radford, Amodei, Christiano, and Irving]{ziegler2019fine}
Daniel~M Ziegler, Nisan Stiennon, Jeffrey Wu, Tom~B Brown, Alec Radford, Dario Amodei, Paul Christiano, and Geoffrey Irving.
\newblock Fine-tuning language models from human preferences.
\newblock \emph{arXiv preprint arXiv:1909.08593}, 2019.

\bibitem[Zollner et~al.(2004)Zollner, Asfour, and Dillmann]{zollner2004programming}
R~Zollner, Tamim Asfour, and R{\"u}diger Dillmann.
\newblock Programming by demonstration: dual-arm manipulation tasks for humanoid robots.
\newblock In \emph{2004 IEEE/RSJ International Conference on Intelligent Robots and Systems (IROS)(IEEE Cat. No. 04CH37566)}, volume~1, pp.\  479--484. IEEE, 2004.

\end{thebibliography}
\bibliographystyle{iclr2024_conference}

\clearpage
\appendix

\section{Action Chunking Technique}
\label{app:ac}
In practice, we find that the errors in action prediction accumulate as the number of historical decisions increases due to the imperfection of the learned policy. This may cause the robot to drift out of the training distribution, reaching hard-to-recover states~\citep{ross2011reduction}. To alleviate this, we prefer to predict multiple actions in one shot, thereby reducing the total number of decisions in a trajectory. In this way, we model $p(\va_{t:t+T_a} | \ell, \vo_t)$, where $\va_{t:t+T_a}:=(\va_t, \dots, \va_{t+T_a-1})$ is an action chunk and $T_a$ denotes the chunk size~\citep{zhao2023learning}. To adapt Eq.~\ref{eq1} and Eq.~\ref{eq2} to this context, we could simply replace $\va_t$ by $\va_{t:t+T_a}$. Besides, according to \citet{chi2023diffusionpolicy}, action chunking is also helpful for improving temporal consistency. It can better consider the coherence of previous and subsequent actions when making decisions and may avoid sudden changes in actions that may cause damage to the robot.

\section{Architecture Details}
\label{app:arc}
\paragraph{Encoding of Multi-Modal Inputs.}
Encoding details are outlined below:
\begin{itemize}[leftmargin=1.0em]
\vspace{-0.5em}
\setlength\itemsep{0em}
    \item \textbf{Low-Dimensional Inputs.} The proprioception $\vz_t$ and the noisy action chunk $\tilde{\va}_{t:t+T_a}$ are first embedded into the unified action space. This space is used to unify the representation of $\vz_t$ and $\tilde{\va}_{t:t+T_a}$ across various robots, which is elaborated in Sec.~\ref{sec:model:data}. Then, they are encoded into the token space by a shared MLP since they have similar physical meanings. Such continuous encoding can avoid precision loss in contrast to discretized encoding~\citep{brohan2022rt,brohan2023rt,kim2024openvla}. For frequency $c$ as well as the diffusion time step $k$, we encode them into the token space through two MLPs, respectively. Afterward, all of them are concatenated together in the length direction to achieve \textit{in-context conditioning}~\citep{peebles2023scalable,bao2023all}, resulting in an input token sequence of length $1+T_a+1+1$. Finally, position embeddings are added to distinguish different modalities and to inject temporal information in $\tilde{\va}_{t:t+T_a}$.
    
    % We adopt \textit{in-context conditioning}~\citep{peebles2023scalable,bao2023all} for low-dimensional conditions, including $\vz_t$, $c$ and $k$, to obtain better conditions injection. To be specific, tokens of noisy inputs $\tilde{\va}_{t:t+T_a}$ and low-dimensional conditions are concatenated together in the length direction, resulting in an input token sequence of length $1+T_a+1+1$. Afterward, position embeddings are added to distinguish different modalities and inject temporal information.
    
    % due to the small number of their tokens. In this way, we concatenate all the tokens (i.e., tokens of $\vz_t$, $\tilde{\va}_{t:t+T_a}$, $c$, and $k$) together in the length direction, resulting in an input token sequence of length $1+T_a+1+1$. We also add position embeddings to it to distinguish different modalities and inject the temporal information into the action chunk $\tilde{\va}_{t:t+T_a}$. 
    \item \textbf{Image Inputs.} We encode the RGB images by a frozen SigLIP~\citep{zhai2023sigmoid} and utilize an additional MLP to project the output to the token space. To enhance the model’s ability to distinguish images based on viewpoint and time steps, we extend traditional sinusoidal positional embeddings to multi-dimensional grids, as shown on the right side of Fig.~\ref{fig:framework}. This modification integrates spatial-temporal information, enabling the model to capture the relationships between input images. Specifically, we adopt the implementation by \citet{liu2022petr}, employing grid dimensions of $(T_\text{img}, N_\text{cam}, N_\text{patch}, D)$. Here, $N_\text{cam}$ represents the number of cameras, set to three in our configuration, and $N_\text{patch}$ indicates the number of patches into which each image is divided by the ViT-based Image Encoder and $D$ denotes the embedding dimension.
    
    % To help the model distinguish images from different viewpoints and time-steps, we further extends traditional sinusoidal positional embeddings to n-dimensional grids of to integrates spatial-temporal information, allowing models to capture the relationships between input images. Concretely, we follow the implementation by \citet{liu2022petr} with the shape of grids is $(T_\text{img}, N_\text{cam}, N_\text{patch}, D)$,
    % where $N_\text{cam}$ denotes number of cameras where $N_\text{cam} = 3$ in our settings, and $N_\text{patch}$ denotes numbers of patches which one image is divided into by the ViT-based Image Encoder and $D$ denotes embedding dimension. 

    \item \textbf{Language Inputs.} Language instruction is encoded by a frozen T5-XXL~\citep{2020t5}, and an MLP is used to project the output to the token space. When calculating attention for language tokens, we apply the language attention mask to mask out the pad tokens appended during batching.
    
    % When performing cross-attention conditioning, we apply the language attention mask to mask out the pad tokens appended during batching.
\end{itemize}
During training, each input from various modalities is independently masked with a probability of $10\%$. 

\paragraph{Network Structure of $f_{\vth}$.}
After encoding, we feed the tokens of the low-dimensional inputs into the main network, which is adjusted from Diffusion Transformers (DiTs) with Cross-Attention~\citep{peebles2023scalable} due to their high scalability. For better training stability, we add QKNorm~\citep{henry2020query} into each attention layer and replace each LayerNorm with RMSNorm~\citep{zhang2019root}. In each DiT block's cross-attention layer, we alternately inject language and image tokens rather than simultaneously inject both, avoiding the issue of token imbalance between the two modalities. After $L$ DiT blocks, we normalize the output and project it back to the action space via an MLP decoder.

\section{Physically Interpretable Unified Action Space}
\label{unified_action_space}
As mentioned in Sec.~\ref{sec:model:data}, we embed the actions of various robots into one unified space that includes all the main physical quantities of robots. This unified action space has a dimensionality of 128. Table~\ref{tab:state_vec_mapping} describes each element of the vector in this unified action space. For a specific robot, each element of the raw action vector is filled into the corresponding position of the unified action vector according to its physical meanings, with the remaining positions being padded. 

% Such mapping preserves the physical meaning of each dimension of the unified action space, facilitating transferring physics knowledge between datasets of different robots. 

% We leave the specific definition of this space in App.~\ref{unified_action_space}.

\begin{table}[ht]
\centering
\begin{tabular}{l c c }
\toprule
\textbf{Index Range} & \textbf{Element Index} & \textbf{Mapped Physical Quantity} \\ \midrule
\texttt{[0, 10)} & 0--9 & Right arm joint positions \\ 
\texttt{[10, 15)} & 10--14 & Right gripper joint positions \\ 
\texttt{[15, 25)} & 15--24 & Right arm joint velocities \\ 
\texttt{[25, 30)} & 25--29 & Right gripper joint velocities \\ 
\texttt{[30, 33)} & 30--32 & Right end effector positions \\ 
\texttt{[33, 39)} & 33--38 & Right end effector 6D pose \\ 
\texttt{[39, 42)} & 39--41 & Right end effector velocities \\ 
\texttt{[42, 45)} & 42--44 & Right end effector angular velocities \\ 
\texttt{[45, 50)} & 45--49 & Reserved \\ 
\texttt{[50, 60)} & 50--59 & Left arm joint positions \\ 
\texttt{[60, 65)} & 60--64 & Left gripper joint positions \\ 
\texttt{[65, 75)} & 65--74 & Left arm joint velocities \\ 
\texttt{[75, 80)} & 75--79 & Left gripper joint velocities \\ 
\texttt{[80, 83)} & 80--82 & Left end effector positions \\ 
\texttt{[83, 89)} & 83--88 & Left end effector 6D pose \\
\texttt{[89, 92)} & 89--91 & Left end effector velocities \\ 
\texttt{[92, 95)} & 92--94 & Left end effector angular velocities \\ 
\texttt{[95, 100)} & 95--99 & Reserved \\ 
\texttt{[100, 102)} & 100--101 & Base linear velocities \\ 
\texttt{[102, 103)} & 102 & Base angular velocities \\ 
\texttt{[103, 128)} & 103--127 & Reserved \\ 
\bottomrule
\end{tabular}
\caption{\textbf{Description of the unified action space vector.} For single-arm robot cases, its arm is mapped to the ``right" arm. For a robot arm with only 6 DoF, its joint positions will be filled in the first $6$ of the $10$ corresponding positions. The same is true for other physical quantities.}
\label{tab:state_vec_mapping}
\end{table}

\section{Pre-Training Datasets}
\label{pretrain_dataset_detail}
Our pre-training dataset collection includes $46$ datasets, with a total scale of $1$M+ trajectories and $21$TB, making it the largest pre-training collection of robotics datasets to date. Table~\ref{tab:dataset_percentages} presents the complete list of our pre-training datasets and their sampling weights. We assign an initial weight of $\sqrt{N_j}$ to each dataset with size $N_j$ and adjust it according to the diversity and quality of each dataset.  Compared to linear weighting, this approach prevents excessive sampling of large datasets while ensuring smaller datasets are adequately sampled, thus enhancing the diversity of pre-training samples in each mini-batch. During the pre-training stage, we further observed and adjusted the weights of different datasets based on their intermediate loss results. We increased the weights of those slow-convergent datasets. 

\paragraph{Main Datasets.} We list some main datasets as follows:  
\begin{itemize}[leftmargin=1.0em]
\vspace{-0.5em}
\setlength\itemsep{0em}
\item \textbf{RT-1 Dataset}~\citep{brohan2022rt} is a large diversve dataset including $130$K trajectories with multiple tasks, objects and environments. It is collected across $13$ different embodiments, each equipping a single exterior RGB camera. The action space includes the 6D end effector (EEF), gripper open, and base displacement with a control frequency of $3$Hz.
\item \textbf{DROID}~\citep{khazatsky2024droid} is a large-scale multi-task dataset with $76$K trajectories and $564$ scenes. It is collected via teleoperating a Franka Panda $7$-DoF Robot Arm, with both wrist and exterior RGB-D cameras. The action space includes $7$-DoF joint positions and a gripper width, while the proprioception additionally includes the 6D EEF with a control frequency of $15$Hz.
% \lsm{Maybe we can replace this one with the factual one? Since it is EEF state space.}
\item \textbf{RH20T}~\citep{fang2023rh20t} is a comprehensive dataset covering 110K trajectories and 140 tasks. It includes four different robotic embodiments and three different camera views, sampled at a frequency of 10Hz. It also includes both long and short tasks. Its state space is a mix of 6-DoF and 7-DoF joint positions, and it features a third-person perspective RGB-D camera.
\item \textbf{Mobile ALOHA Dataset}~\citep{fu2024mobile} is a bimanual dataset containing 1K+ trajectories collected by the Mobile ALOHA robot. Its state space includes base movements and 14-dimensional joint positions of both hands, along with three or four first-person perspective cameras. Some of its data includes wide-ranging perspective changes and base movements, which were originally suitable for imitation learning. 
% \lsm{Change with another example...}
\item \textbf{Other Datasets.} The other data come from RH20T~\citep{fang2023rh20t}, RoboSet~\citep{RoboHive}, BridgeData V2~\citep{walke2023bridgedata}, and Open X-Embodiment~\citep{padalkar2023open}. Most of them feature different robotic morphology and camera observation, enhancing both heterogeneity and variety of our pretraining datasets.

% of the action space 
% embodiments and greatly vary in the action space, camera views, kinematic configuration, end-effector morphology, and sampling frequencies, enhancing the heterogeneity and variety of our pretraining dataset.
\end{itemize}

\paragraph{Data Cleaning.} 
Repetitive episodes and episodes of failure are excluded to ensure the quality of the pre-training datasets. We remove blank images, exclude erroneously recorded velocities, and filter out overly short trajectories. Overlength trajectories will be downsampled to avoid unfairness.

\paragraph{Preprocessing of Multi-Modal Observation/Action Inputs.} We describe the preprocessing details of each modality:
\begin{itemize}[leftmargin=1.5em]
\vspace{-0.5em}
\setlength\itemsep{0em}
    \item \textbf{Language Instruction $\ell$.} We perform a simple cleaning on the raw text, such as removing illegal characters and extra spaces, capitalizing the beginning of sentences, and adding a period at the end of sentences. We leave the text variable-length.  
    \item \textbf{RGB Images $\mX_{t-T_{\text{img}}+1:t+1}$.} We employ a fixed-length image input strategy. We fix the image input order and format for all robots, with a total of three views: a static exterior view, a right-wrist view, and a left-wrist view, deemed sufficient for the requirements of most bimanual tasks. We treat a single-arm robot's wrist camera as the right-wrist one and pad the unavailable views with the background color. When fed into the model, each image is padded into a square and resized to $384\times 384$, keeping its origin aspect ratio. Besides, we choose $T_{\text{img}}=2$ since a history length of two is adequate for most situations, striking a balance between efficiency and performance \citep{team2023octo,wu2024embodied}. Finally, we can write the image inputs as $\mX_{t-1:t+1}:=(\{ \mX^1_{t-1}, \mX^2_{t-1},\mX^3_{t-1}\}, \{ \mX^1_{t}, \mX^2_{t},\mX^3_{t}\})$.

    % We fix the number and input order of the images.  For unavailable images, we pad them with the background color. 
    
    \item \textbf{Proprioception $\vz_t$ and Action Chunk $\va_{t:t+T_a}$.} We roughly align the scales of various datasets by unifying the units of physical quantities (m, rad, m/s, rad/s, etc) rather than strictly normalizing to $[-1,1]$ or $\mathcal{N}(0, 1)$ as in prior work~\citep{chi2023diffusionpolicy,team2023octo}. For example, ``$1$ (m)" in different datasets corresponds to the same real-world length. Rescaling the physical quantities will destroy such shared properties and thus impair the model's ability to transfer across robots. We also employ the 6D representation~\citep{8953486} for the EEF rotation to overcome the gimbal lock issue.
    
    % Given that robotic data have physical meanings. We argue that maintaining the scale information is essential for learning the cross-robot physical features. 

    % Small datasets to bad statistics bias? 

    Before choosing $T_a=64$, we have referred to the previous ablation studies by~\citet{zhao2023learning} and balanced between the performance and computational overhead. Besides, historical proprioceptions $\vz_{i}, i<t$ are excluded to prevent the model from learning shortcuts using the low-dimensional inputs only and thus sticking to fixed motion patterns. Instead, we encourage the model to learn generalizable decision-making structures from high-dimensional image features.

    \item \textbf{Control Frequency $c$.} In addressing the challenge posed by differing control frequencies across datasets, we feed the control frequency into the model, allowing the model to take this variation into account when making decisions.
\end{itemize}

\begin{table}[!ht]
\centering
\begin{tabular}{lc}
\toprule
Pre-Training Dataset & Sample Percentage (\%) \\
\midrule
RT-1 Dataset~\citep{brohan2022rt} & 9.00 \\
TACO Dataset~\citep{rosete2022tacorl} & 1.99 \\
JACO Play Dataset~\citep{dass2023jacoplay} & 1.10 \\
Cable Routing Dataset~\citep{luo2023multistage} & 0.27 \\
NYU Door Opening~\citep{pari2021surprising} & 0.33 \\
Viola~\citep{zhu2022viola} & 0.40 \\
Berkeley UR5~\citep{BerkeleyUR5Website} & 1.06 \\
TOTO~\citep{zhou2023train} & 1.06 \\
Kuka~\citep{kalashnikov2018qt} & 1.66 \\
Language Table~\citep{lynch2022interactivelanguagetalkingrobots} & 3.32 \\
Columbia Cairlab Pusht Real~\citep{chi2023diffusionpolicy} & 0.40 \\
Stanford Kuka Multimodal Dataset~\citep{lee2019icra} & 1.83 \\
Stanford Hydra Dataset ~\citep{belkhale2023hydra} & 0.80 \\
Austin Buds Dataset~\citep{zhu2022bottom} & 0.23 \\
Maniskill Dataset~\citep{gu2023maniskill2} & 5.78 \\
Furniture Bench Dataset~\citep{heo2023furniturebench} & 2.36 \\
UCSD Kitchen Dataset~\citep{ucsd_kitchens} & 0.40 \\
UCSD Pick And Place Dataset~\citep{Feng2023Finetuning} & 1.23 \\
Austin Sailor Dataset~\citep{nasiriany2022sailor} & 0.50 \\
Austin Sirius Dataset~\citep{liu2022robot} & 0.80 \\
BC Z~\citep{jang2021bc} & 6.91 \\
UTokyo PR2 Opening Fridge~\citep{oh2023pr2utokyodatasets} & 0.30 \\
UTokyo PR2 Tabletop Manipulation~\citep{oh2023pr2utokyodatasets} & 0.50 \\
UTokyo Xarm Pick And Place~\citep{matsushima2023weblab} & 0.33 \\
UTokyo Xarm Bimanual~\citep{matsushima2023weblab} & 0.03 \\
Berkeley MVP~\citep{Radosavovic2022} & 0.73 \\
Berkeley RPT~\citep{Radosavovic2022} & 1.00 \\
KAIST Nonprehensile~\citep{kimpre} & 0.46 \\
Tokyo U LSMO~\citep{Osa22} & 0.23 \\
DLR Sara Grid Clamp~\citep{padalkar2023guided} & 0.03 \\
Robocook~\citep{shi2023robocook} & 1.66 \\
Imperialcollege Sawyer Wrist Cam~\citep{imperialcollege_sawyer_wrist_cam} & 0.43 \\
Iamlab CMU Pickup Insert~\citep{saxena2023multiresolution} & 0.83 \\
UTAustin Mutex~\citep{shah2023mutex} & 1.29 \\
Fanuc Manipulation~\citep{fanuc_manipulation2023} & 0.66 \\
Play Fusion~\citep{chen2023playfusion} & 0.80 \\
DROID~\citep{khazatsky2024droid} & 10.06 \\
FMB~\citep{luo2024fmbfunctionalmanipulationbenchmark}& 1.39 \\
Dobb·E~\citep{shafiullah2023bringing} & 1.20 \\
QUT Dexterous Manipulation~\citep{ceola2023lhmanip} & 0.46 \\
Aloha Dataset~\citep{zhao2023learning} & 4.98 \\
Mobile Aloha Dataset~\citep{fu2024mobile} & 4.98 \\
RoboSet~\citep{RoboHive} & 4.48 \\
RH20T~\citep{fang2023rh20t} & 10.99 \\
Calvin Dataset~\citep{mees2022calvin} & 3.32 \\
BridgeData V2~\citep{walke2023bridgedata} & 7.44 \\
\bottomrule
\end{tabular}
\caption{The pre-training datasets and their corresponding weights.}
\label{tab:dataset_percentages}
\end{table}

\section{Fine-Tuning Dataset}
\label{app:ft_dataset}

\begin{figure}[!tb]
    \centering
    \includegraphics[width=\textwidth]{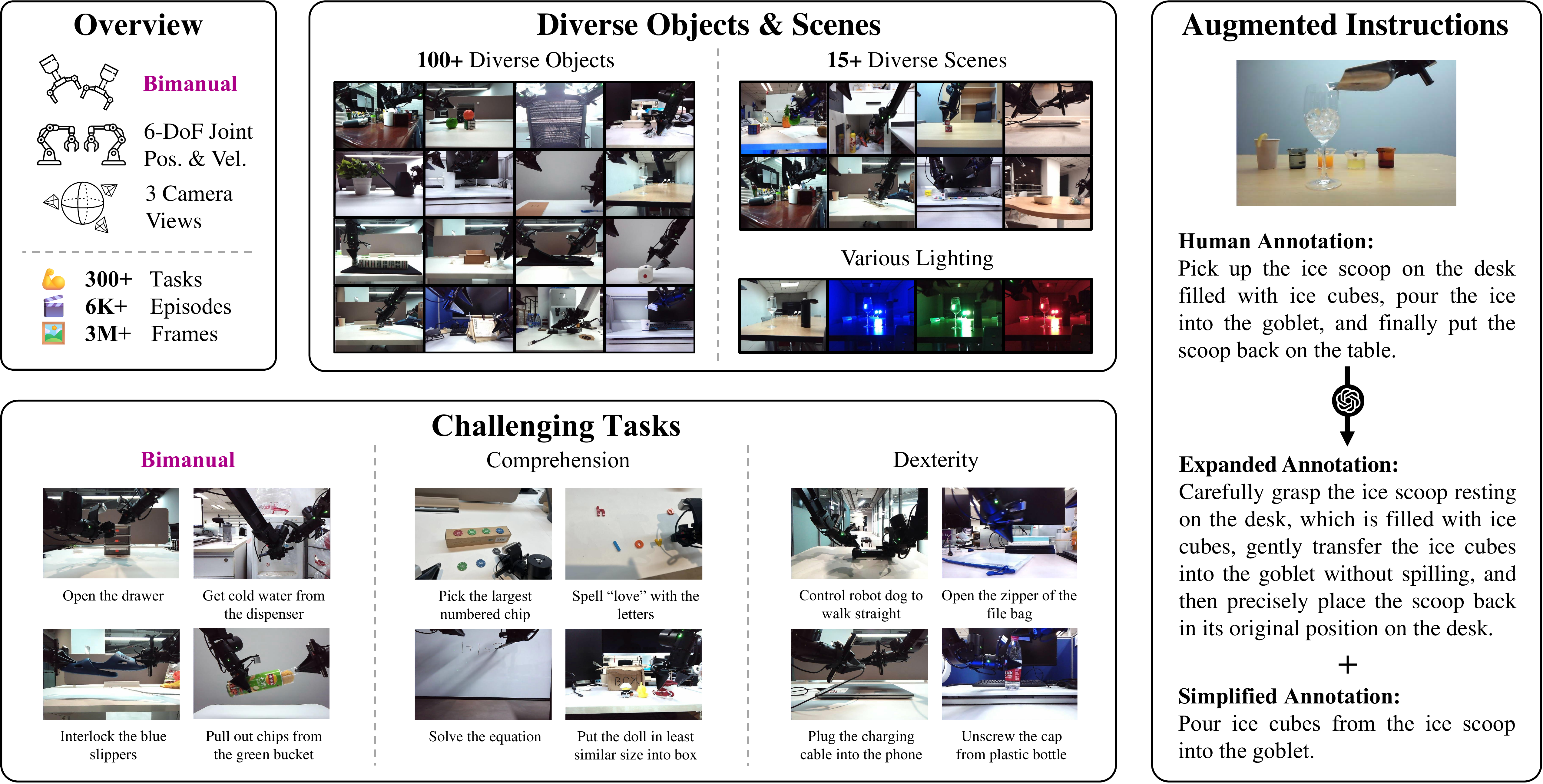}
    \caption{\textbf{Fine-Tuning dataset.} Our dataset includes the following key features: (1) \textbf{Diverse Objects and Scenes.} Our dataset contains objects with different properties manipulated in different scenes and conditions. (2) \textbf{Challenging Tasks.} Our dataset incorporates dexterous manipulation, language and vision comprehension, and bimanual tasks. (3) \textbf{Multi-Modal Features.} Our dataset is annotated with rich multi-modal data, including 3-View RGB cameras, joint information, and augmented instructions.}
    \label{fig:ft_dataset}
\end{figure}

Our fine-tuning dataset is created using Mobile ALOHA robot~\citep{fu2024mobile}, including $300$+ tasks, $6$K+ trajectories, and $3$M+ frames. It is also one of the largest open-source multi-task bimanual robot datasets to date. Fig.~\ref{fig:ft_dataset} gives a summary of this dataset. We have borrowed $3$ tasks ($140$ episodes in total) from the open-source \href{https://openi.pcl.ac.cn/ARIO/Songling_datasets/datasets}{Songling dataset}~\citep{wang2024all}.
% \vspace{-0.5em}
\begin{itemize}[leftmargin=1.0em]
\vspace{-0.5em}
\setlength\itemsep{0em}
\item \textbf{Multi-Modal Features.} We collect the dataset with three RGB cameras positioned at the front and on the left and right grippers. We record dual-arm 6-DoF joint positions and velocities, along with the gripper angles. We manually annotated instructions for each task. To further augment our instructions and align them with the pre-training datasets, we utilize GPT-4-Turbo~\citep{achiam2023gpt} to generate $100$ expanded instructions and one simplified instruction for each task. This multi-modal information further enhances the richness and quality of our dataset.
\item  \textbf{Diverse Objects and Scenes.} Our dataset includes diverse tasks and scenes, encompassing more than 300 tasks, including skills such as picking up, inserting, writing, pushing, and pulling. It features 100+ objects with rigid and non-rigid bodies of various sizes and textures. We collect the dataset in 15+ scenes and introduce randomness during data collection for each task, such as varying the initial positions of objects and robots. To further increase diversity, we added random lighting conditions. For instance, pouring water was performed under both normal lighting and changing color conditions. These measures further enhance the diversity of our dataset. 
\item \textbf{Challenging Tasks.} Various challenging tasks are also considered, encompassing dexterous manipulations, such as unscrewing the cap from a plastic bottle, and comprehension tasks, such as spelling ``love'' with letter blocks. Furthermore, the dataset includes tasks that integrate both dexterity and comprehension, such as solving mathematical equations on the whiteboard. Additionally, our dataset incorporates bimanual tasks, such as inserting the charging cable into the phone. These complex, high-quality tasks further enhance the model's downstream comprehensibility and generalizability.
\end{itemize}

\section{RDT Training Details}
\label{app:train}

\paragraph{Platform.} We use Pytorch~\citep{paszke2019pytorch} and DeepSpeed~\citep{rasley2020deepspeed} to facilitate parallel training and employ a producer-consumer framework with TensorFlow Dataset~\citep{TFDS} for fast data loading. Since most of the datasets in the Open X-Embodiment~\citep{padalkar2023open} are stored in the form of \texttt{TFRecord}, we convert all pre-training datasets into \texttt{TFRecord} for storage. In pre-training, we use the producer process to decompress the data from \texttt{TFRecord} and store it in a buffer on the hard disk. At the same time, we use the consumer process to read data from the buffer in a disorderly order and feed it to the model training. This not only decouples the TensorFlow~\citep{tensorflow2015-whitepaper} and PyTorch environments but also alleviates the training performance loss caused by the small size of the shuffling buffer in the memory. In the fine-tuning stage, since the dataset is relatively small, we additionally implement a data reading pipeline using the HDF5 dataset for storage.

\paragraph{Padding Action and Proprioception.}
To embed a specific robot action into the $128$-dimensional unified action space, we need to pad unavailable action elements. The usual practice is to pad with a 0 value or a specific value. But ``0" actually has a physical meaning. For example, a speed of ``0" generally represents stillness relative to the ground. This may confuse the model: Does ``0" represent stillness or a filler value? To solve this problem, we concatenate the action and proprioception with a $0$-$1$ vector indicating whether each dimension is padded before encoding them into the token space, resulting in a $256$-dimensional vector. This can supplement the missing availability information and eliminate confusion.

\paragraph{Inspecting Training Process.}
During training, for every fixed period, we conduct a diffusion sampling and compare the sampled actions with the ground truth of the training dataset. Empirically, we discover a general positive correlation between the Mean Squared Error (MSE) of the two and the performance of deployment on the robot. This observation allows us to monitor the model's training progress easily. When this MSE converges, we can generally stop training. We note that an overly low MSE may also mean overfitting.

% \paragraph{Some Experimental Details.}
% remove the first still steps
% equally sample from instruction, simplified instruction, and expanded instructions.
% no cfg

\paragraph{Data Augmentation.}
Overfitting is a common challenge in training large neural models, particularly in the fine-tuning phase. We utilize data augmentation techniques to resolve it. We perform image augmentation, including color jittering and image corruption, and add Gaussian noise to the input proprioception with a signal-to-noise ratio (SNR) of $40$dB. We also use GPT-4-Turbo to augment and expand the language instructions (Refer to Sec.~\ref{app:ft_dataset} for more details on the instruction augmentation). 

\paragraph{Some Fine-Tuning Details.}
During fine-tuning, we removed a static part at the beginning of each episode, which might be caused by the operator not reacting after the recording started. Our language instructions are sampled from the original manually annotated instruction, the expanded instructions, and the simplified instruction with a probability of one-third. When the expanded instructions are drawn, we evenly sample one from the $100$ expanded instructions corresponding to the task. We did not apply Classifier-Free Guidance (CFG) because we found that this did not improve the performance of the model but instead brought the unstable robot arm behavior.

\section{Hardware Details}
\label{hardware_detail}

\begin{figure}[!ht]
    \centering
    \begin{minipage}[b]{0.4\textwidth} 
        \centering
        \includegraphics[width=\textwidth]{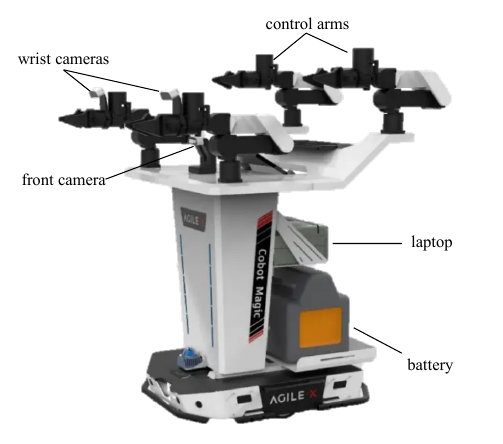}
        \caption{Hardware features.}
        \label{fig:aloha}
    \end{minipage}
    \hfill
    \begin{minipage}[b]{0.55\textwidth}
        \centering
        \begin{tabular}{l c}
            \toprule
            Parameter & Value \\
            \midrule
            DoF & 7 $\times$ 2 = 14 \\
            Size & 1080 $\times$ 700 $\times$ 1140 \\
            Arm weight & 4.2~kg \\ 
            Arm Payload & 3000~g (peak) \\
             & 1500~g (valid) \\
            Arm reach & 600~mm \\
            Arm repeatability & 1~mm \\
            Arm working radius &  653 mm \\ 
            Joint motion range & J1: ±154°, J2: 0°$ \sim $165° \\
            & J3: -175°$ \sim $0°, J4:  ±106°\\
            & J5: ±75° , J6: ±100° \\
            Gripper range & 0-80~mm \\
            Gripper max force & 10~NM \\
            \bottomrule
        \end{tabular}
        \captionof{table}{Technical specifications.}
        \label{tab:tech_spec}
    \end{minipage}
\end{figure}

We provide a detailed overview of the hardware configuration of our target dual-arm robot. Our model is deployed and evaluated on the Cobot Mobile ALOHA, a robot using the Mobile ALOHA system design~\citep{fu2024mobile} and manufactured by \href{https://global.agilex.ai}{agilex.ai}. The key features of the robot are illustrated in Fig.~\ref{fig:aloha}
. It is equipped with two wrist cameras, a front camera, a laptop, and an onboard battery. The robot's technical specifications are listed in Table~\ref{tab:tech_spec}. It is important to note we used the ``mobile" ALOHA only to facilitate transportation and testing between various scenes and did not use its autonomous mobility feature during any training or inference stages. Our tasks are still static bimanual manipulation tasks. 

 % that the robot's tracer was not utilized during any training or inference stages.

\begin{table}[ht]
\caption{\textbf{Comparision of different baselines.} We compare baselines as well as different variants of our model in terms of model size, data size, and modeling scheme.}
\label{tbl:baseline}
\renewcommand{\arraystretch}{1.2}
\begin{center}
\resizebox{\linewidth}{!}{
\begin{tabular}{lccc}
\toprule
\multicolumn{1}{c}{\textbf{METHOD NAME}} & \multicolumn{1}{c}{\textbf{LARGE MODEL}}             & \multicolumn{1}{c}{\textbf{LARGE MULTI-ROBOT DATA}}   &  \multicolumn{1}{c}{\textbf{MODELING}}                                               \\ \hline
ACT~\citep{zhao2023learning} & \ding{55} & \ding{55} & VAE \\
OpenVLA~\citep{kim2024openvla} & \ding{51} & \ding{51} & Discretization \\
Octo~\citep{team2023octo} & \ding{55} & \ding{51} & Diffusion \\\hline
RDT (scratch) & \ding{51} & \ding{55} & Diffusion \\
RDT (small) & \ding{55} & \ding{51} & Diffusion \\
RDT (regress) & \ding{51} & \ding{51} & Regression \\
RDT (\textbf{ours}) & \ding{51} & \ding{51} & Diffusion
\\ \bottomrule
\end{tabular}
}
\end{center}
\end{table}

\section{Experiment Details}
\label{app:exp}

\paragraph{Calculation of Total Performance.} The general performance in Fig.~\ref{fig:head-demo} of each method is calculated in three steps. Firstly, we calculate the success rate of a method in each task. We take an average of the total success rate and any additional requirement, i.e., the average of the values in the \textit{Total} column and all columns to its right in Table~\ref{tbl:result}. For example, in the \textit{Pour Water-L-1/3}, we take the average of \textit{Total}, \textit{Correct Hand}, and \textit{Correct Amount}. Secondly, we calculate the success rate of each dimension of \textit{Unseen Object}, \textit{Unseen Scene}, \textit{Instruction Following}, \textit{Few-Shot Learning}, and \textit{Dexterity} by averaging all the tasks in this dimension (see Table~\ref{tbl:taskdim} for the correspondence). Lastly, we average the success rates of all the dimensions to obtain the overall result.

\paragraph{Implementation and Hyper-Parameters of RDT.}
We list the details of the multi-modal encoders in Table~\ref{tab:encoder_configs} and the model parameter in Table~\ref{tab:model_configs}. The image history size is $T_{\text{img}}=2$, the action chunk size is $T_a = 64$, the language token space dimension is $4096$, the image token space dimension is $1152$, and the token space dimension of RDT is $2048$. We use adaptors to align each modality's token dimension to $2048$. And all adaptors for multi-modal encoders are with GeLU activation~\citep{hendrycks2016gaussian}.

We use the AdamW optimizer~\citep{adam2019no} with a constant learning rate scheduler and hyper-parameters in Table~\ref{tab:rdt_hyper_params} in the pre-training and fine-tuning stages. The model is pre-trained and finetuned on $48$ H100 80GB GPUs for $1$M steps and $130$K steps, respectively. Due to scheduling reasons, we did not start fine-tuning from the 1M pre-trained checkpoint but chose the 500K checkpoint. During the training stage, we use the DDPM scheduler with a glide cosine scheduler (i.e., \texttt{squaredcos\_cap\_v2}) and a step number of $1000$. During the sampling stage, we utilize the DPM-Solver++~\citep{lu2022dpm} with a glide cosine scheduler and a sampling step number of $5$. During fine-tuning, we also filter out episodes with a length lower than $32$ and down-sample those with a length higher than $2048$ to $2048$.

\begin{table}[ht]
\centering
\begin{tabular}{lccc}
\toprule
Modality & Encoder & Trainable & Adaptor \\
\midrule
Language & T5-XXL~\citep{2020t5} & N & 2-layers MLP\\
Image & SigLIP~\citep{zhai2023sigmoid} & N & 2-layers MLP\\
Action & - & - & 3-layers MLP \\
\bottomrule
\end{tabular}
\caption{Encoder configurations of RDT.}
\label{tab:encoder_configs}
\end{table}

\begin{table}[ht]
\centering
\begin{tabular}{lccccc}
\toprule
Model & Layers & Hidden size  & Heads & \#Params \\
\midrule
RDT-1B & 28 & 2048 & 32 & 1.2B \\
\bottomrule
\end{tabular}
\caption{Model configurations for RDT.}
\label{tab:model_configs}
\end{table}

\begin{table}[ht]
    \centering
    \begin{tabular}{l c}
         \toprule
         \textbf{Hyper-Parameter} & \textbf{Value} \\
         \midrule
         Batch Size & 32$\times$48 \\
         Learning Rate &  $1\times10^{-4}$\\
         Mixed Precision & \texttt{bf16} \\
         Warm-Up Steps & $500$ \\
         $\beta_1$ & $0.9$ \\
         $\beta_2$ & $0.999$ \\
         Weight Decay & $1\times10^{-2}$ \\
         $\epsilon$ & $1\times10^{-8}$ \\
        \bottomrule
    \end{tabular}
    \caption{Hyper-parameters for both pre-training and fine-tuning RDT.}
    \label{tab:rdt_hyper_params}
\end{table}

\paragraph{Implementation and Hyper-Parameters of ACT.} 
We directly employed the same architecture and hyper-parameters of ACT as that in the original paper~\citep{fu2024mobile}, except for the hyper-parameters in Table~\ref{tab:act_hyper_params}. We trained ACT with $90\%$ of the $6$K fine-tuning episodes for $8000$ epochs (about $8$ days in total), while the remaining $10\%$ is treated as the validation set. We took the checkpoint at epoch $5413$ as the final outcome, according to the best performance in the validation set.

\begin{table}[ht]
    \centering
    \begin{tabular}{l c}
         \toprule
         \textbf{Hyper-Parameter} & \textbf{Value} \\
         \midrule
         Batch Size & 80$\times$4 \\
         Learning Rate &  $9\times10^{-5}$\\
         Learning Rate for Backbone&  $4\times10^{-5}$\\
        \bottomrule
    \end{tabular}
    \caption{Adapetd hyper-parameters of ACT.}
    \label{tab:act_hyper_params}
\end{table}

\begin{figure}[ht]
    \centering
    \includegraphics[width=\linewidth]{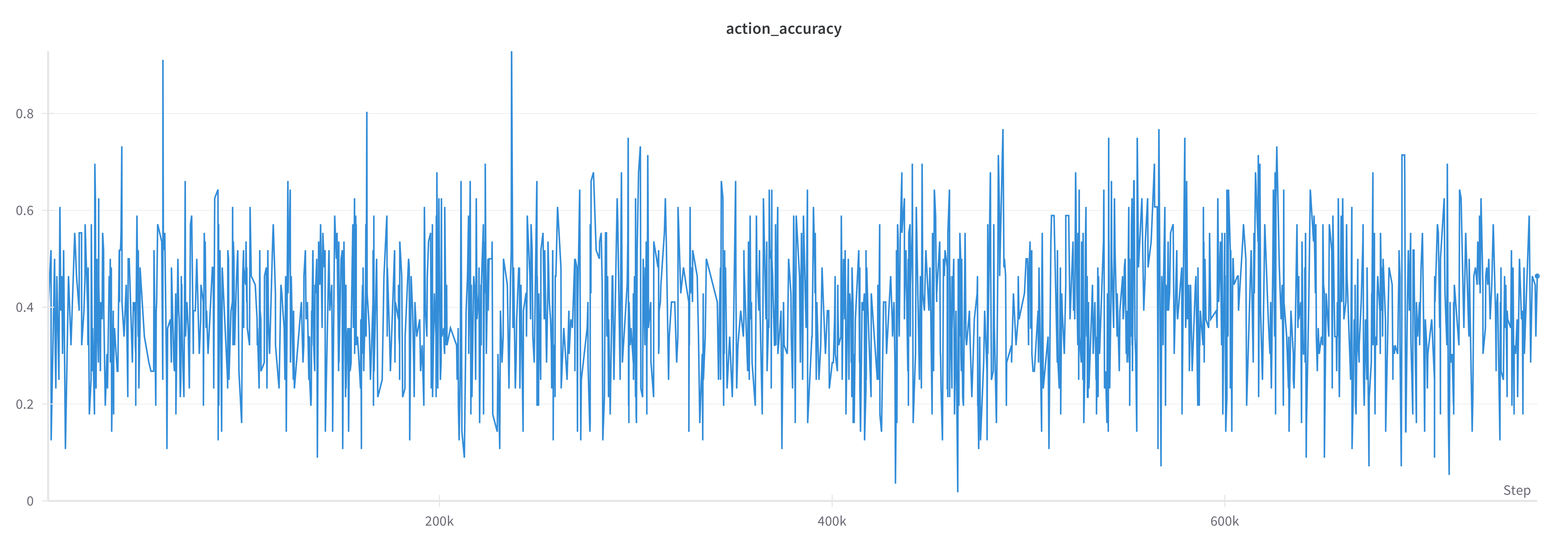}
    \caption{The accuracy of action token prediction fluctuates rather than converges with the number of training steps when fine-tuning OpenVLA with the full fine-tuning dataset.}
    \label{fig:train_openvla}
\end{figure}

\paragraph{Implementation and Hyper-Parameters of OpenVLA.} We adopt the official implementation (\url{https://github.com/openvla/openvla}) and flagship pre-trained model and checkpoint at \url{https://huggingface.co/openvla/openvla-7b}. For each task in evaluation, we further fine-tune the officially pre-trained OpenVLA with all the task-relevant demonstrations ($\sim100$ episodes) from the fine-tuning dataset to facilitate convergence and train the model to around $95\%$ action token accuracy as suggested by~\citet{kim2024openvla} (\url{https://github.com/openvla/openvla/issues/12#issuecomment-2203772810}). Additionally, we experimented with both full-parameter tuning and LoRA methods using the entire dataset but did not achieve sufficient action token accuracy (approximately $60\%$) for deployment upon convergence (see Fig.~\ref{fig:train_openvla}). According to real-robot testing, such non-convergent checkpoints exhibit completely static or random 
 behaviors in the deployment.

Concretely, we adhere to the same hyper-parameters claimed in \citet{kim2024openvla} for fine-tuning via LoRA~\citep{hu2021lora} as detailed in Table~\ref{tab:openvla_hyper_params}. 

\begin{table}[ht]
    \centering
    \begin{tabular}{l c}
          \toprule
         \textbf{Hyper-Parameter} & \textbf{Value} \\
         \midrule
         Batch Size & 16$\times$8 \\
         Learning Rate &  $2\times10^{-5}$\\
         Lora Rank & 32 \\
         Image Augmentation & \texttt{True} \\
        \bottomrule
    \end{tabular}
    \caption{Hyper-parameters of fine-tuning OpenVLA for bimanual manipulations.}
     \label{tab:openvla_hyper_params}
\end{table}

\paragraph{Implementation and Hyper-Parameters of Octo.} We utilize the official implementation available at \url{https://github.com/octo-models/octo} and the most comprehensive pre-trained model, \texttt{octo-base-1.5}, hosted at \url{https://huggingface.co/rail-berkeley/octo-base-1.5}. We follow the officially recommended practices for fine-tuning a bimanual robot, detailed in \url{https://github.com/octo-models/octo/blob/main/examples/02_finetune_new_observation_action.py}, employing a full-parameter approach. Additionally, we have incorporated an extra image tokenizer to process images from the right-wrist camera, enhancing the system’s manipulation capabilities. Furthermore, by integrating image augmentation during the fine-tuning process, we enhance the performance upon deployment in real-world robotics. We replicate the wrist image tokenizer from the pre-trained model to initialize the right-wrist image tokenizer. Similar to OpenVLA, we only fine-tune octo with the task-relevant demonstrations for each evaluation tasks, for we do not observe sufficient test MSE (approximately $10^{-1}$) for deployment upon convergence (Fig.~\ref{fig:train_octo}). Concretely, we apply the default hyper-parameters with variations listed in Table~\ref{tab:octo_hyper_params}:

\begin{figure}[ht]
    \centering
    \includegraphics[width=\linewidth]{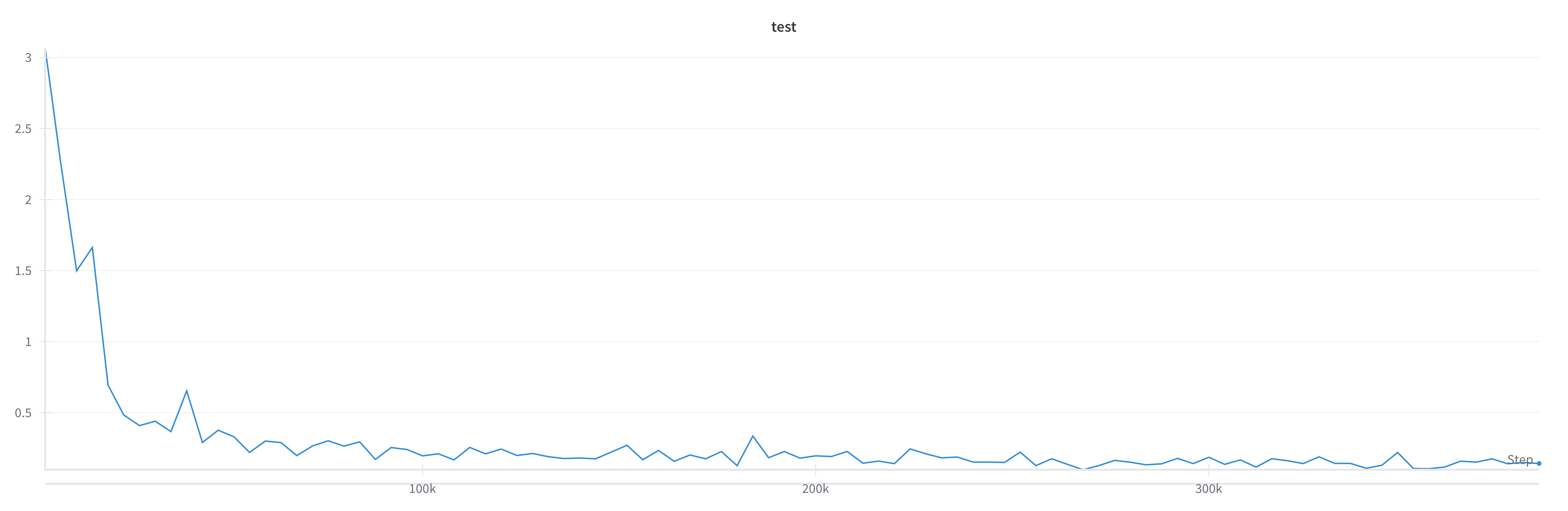}
    \caption{The test MSE of action prediction fluctuates rather than converges with the number of training steps when fine-tuning Octo with the full fine-tuning dataset.}
    \label{fig:train_octo}
\end{figure}oct

\begin{table}[ht]
    \centering
    \begin{tabular}{l c}
          \toprule
         \textbf{Hyper-Parameter} & \textbf{Value} \\
         \midrule
         Action Head Type & \texttt{DiffusionActionHead} \\
         Batch Size & 8$\times$8 \\
         Action Chunk Size & 8 \\
         \multirow{ 4}{*}{Image Augmentation} & \texttt{RandomBrightness(0.1)}  \\
                                              & \texttt{RandomContrast(0.9, 1.1)}  \\
                                              & \texttt{RandomSaturation(0.9, 1.1)}  \\
                                              & \texttt{RandomHue(0.05)}  \\
        \bottomrule
    \end{tabular}
    \caption{Hyper-parameters of fine-tuning Octo for bimanual manipulations.}
     \label{tab:octo_hyper_params}
\end{table}

% For each task in evaluation, we further fine-tune the officially pre-trained OpenVLA with all the task-relevant demonstrations ($\sim100$ episodes) from the fine-tuning dataset to facilitate convergence and train the model to around $95\%$ action token accuracy as suggested by~\citet{kim2024openvla} (\url{https://github.com/openvla/openvla/issues/12#issuecomment-2203772810}). Additionally, we experimented with both full-parameter tuning and LoRA methods using the entire dataset but did not achieve sufficient action token accuracy (approximately $60\%$) for deployment upon convergence (see Fig.~\ref{fig:train_openvla}). According to real-robot testing, such non-convergent checkpoints exhibit completely static or random 
%  behaviors in the deployment.

% \lsmtodo{Hyper-Parameter of model, training, why not use full data, data used for each task}

\section{More Results}

\begin{figure}[ht]
    \centering
    \includegraphics[width=\linewidth]{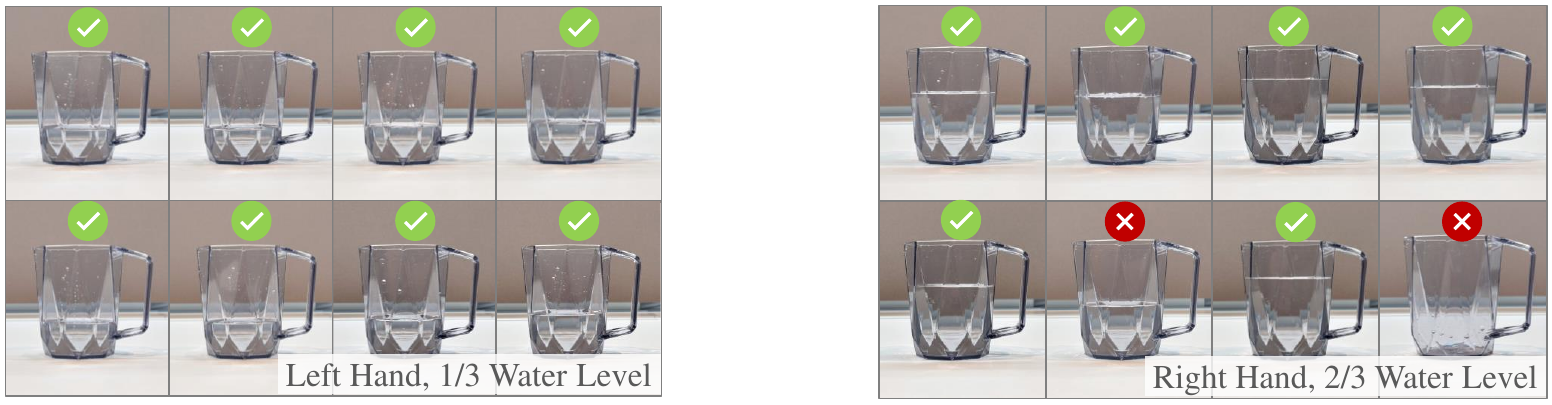}
    \caption{Visulization of the resulting water levels across $8$ trails in \textit{Pour Water-L-1/3} and \textit{Pour Water-R-2/3}. \textbf{Left:} The water level completed by RDT in each trial is extremely close to the ground-truth 1/3 standard. \textbf{Right:} RDT made one mistake in pouring (empty cup) and one mistake in water level, but the other trials were in roughly good agreement with 2/3.
    }
    \label{fig:water}
\end{figure}

We further provide a zoom-in view for water-level across $8$ trails in instruction-following evaluation in Fig.~\ref{fig:water}. 

% \section{More Results}
% \label{more_results}
% In this section, we demonstrate more experimental results with RDT-1B. More results can be found on our project website.
% \end{document}

\end{document}